\documentclass[runningheads]{llncs}
\usepackage{graphicx}
\usepackage{comment}
\usepackage{amsmath,amssymb} % define this before the line numbering.
\usepackage{color}
\usepackage[width=122mm,left=12mm,paperwidth=146mm,height=193mm,top=12mm,paperheight=217mm]{geometry}
\usepackage[pagebackref=true,breaklinks=true,letterpaper=true,colorlinks,bookmarks=false]{hyperref}

\graphicspath{{./figures/}}
\usepackage{parskip}
\def\ie{\emph{i.e.\ }} 
\def\eg{\emph{e.g.\ }}

\begin{document}
% \linenumbers
\pagestyle{headings}
\mainmatter
\def\ECCVSubNumber{2874}  % Insert your submission number here

\title{Meta-Sim2: Unsupervised Learning of Scene Structure for Synthetic Data Generation} % Replace with your title
\begin{comment}
\titlerunning{ECCV-20 submission ID \ECCVSubNumber} 
\authorrunning{ECCV-20 submission ID \ECCVSubNumber} 
\author{Anonymous ECCV submission}
\institute{Paper ID \ECCVSubNumber}
\end{comment}
%******************

% \begin{comment}
\titlerunning{MetaSim2: Learning to Generate Synthetic Scene Structures}
\author{Jeevan Devaranjan\thanks{authors contributed equally, work done during JD's internship at NVIDIA}$^{1,3}$ \and
Amlan Kar$^{\star 1,2,4}$ \and
Sanja Fidler$^{1,2,4}$}
\authorrunning{Devaranjan, Kar et al.}
\institute{$^1$NVIDIA \hspace{1.5mm} $^2$University of Toronto
\hspace{1.5mm} $^3$University of Waterloo \hspace{1.5mm} $^4$Vector Institute}
% \end{comment}
%******************
\maketitle
\begin{abstract}

Procedural models are being widely used to synthesize scenes for graphics, gaming, and to create (labeled) synthetic datasets for ML. In order to produce realistic and diverse scenes, a number of parameters governing the procedural models have to be carefully tuned by experts. These parameters control both the \emph{structure} of scenes being generated (\eg how many cars in the scene), as well as \emph{parameters} which place objects in valid configurations. Meta-Sim aimed at automatically tuning parameters  given a target collection of real images in an unsupervised way. In Meta-Sim2, we aim to learn the scene \emph{structure} in addition to parameters, which is a challenging problem due to its discrete nature. Meta-Sim2 proceeds by learning to sequentially sample rule expansions from a given probabilistic scene grammar. Due to the discrete nature of the problem, we use Reinforcement Learning to train our model, and design a feature space divergence between our synthesized and target images that is key to successful training. Experiments on a real driving dataset show that, without any supervision, we can successfully learn to generate data that captures discrete structural statistics of objects, such as their frequency, in real images. We also show that this leads to downstream improvement in the performance of an object detector trained on our generated dataset as opposed to other baseline simulation methods. Project page: \url{https://nv-tlabs.github.io/meta-sim-structure/}. 
\iffalse
OLD
  Procedural models are being widely used to synthesize scenes for graphics, gaming and to create synthetic datasets for ML. Their sampling procedures typically require experts to specify certain structural aspects such as scene layout information, which are hard to design since real scenes that they aim to emulate are highly complex and diverse. We propose a procedural generative model of scenes, which generates an abstract scene representation (scene graphs) that can be rendered into synthetic (labelled) images. It does so by learning to sequentially sample rule expansions from a given probabilistic scene grammar. Our method operates completely unsupervised, \ie without any ground truth scene structure annotations on real data, using a feature space divergence between our synthesized and target images. Experiments on two synthetic datasets and on a real driving dataset show that we can successfully learn to generate data that captures discrete structural statistics of objects, such as their frequency, in real images \emph{unsupervised}. We also show that this leads to downstream improvement in the performance of an object detector trained on our generated data as opposed to other baseline simulation methods. More details can be found at: \url{https://nv-tlabs.github.io/meta-sim-structure/}
  \fi
\end{abstract}

%%%%%%%%% BODY TEXT
\section{Introduction}
\label{introduction}
% \SF{general comment is that there is a lot of italics which is somewhat distracting}
%Synthetic datasets such as Virtual KITTI~\cite{}, GTA~\cite{} and SDR~\cite{} have 
Synthetic datasets are creating an appealing opportunity for training machine learning models \eg for perception and planning in driving~\cite{RosCVPR16,VirtualKITTI,Richter16}, indoor scene perception~\cite{mccormac2016scenenet,savva2019habitat}, and robotic control~\cite{deepmindcontrolsuite2018}.
% since they offer a plethora of labels that can be obtained through a graphics engine. Furthermore, they provide perfect ground-truth for tasks in which it is either expensive or even impossible to obtain such as segmentation, optical flow, depth or material information. Adding a new type of label to synthetic datasets is as simple as calling a graphics renderer, rather than embarking on a time consuming annotation endeavor which requires developing new tooling, hiring and training annotators, and overseeing their work. 
Via graphics engines, synthetic datasets come with perfect ground-truth for tasks in which labels are expensive or even impossible to obtain, such as segmentation, depth or material information. Adding a new type of label to synthetic datasets is as simple as calling a renderer, rather than embarking on a time consuming annotation endeavor that requires new tooling and hiring, training and overseeing annotators.

%With learning, we can enable data-dependent scenario simulation. This allows targeted synthetic data creation for machine learning and robotics (such as for self-driving cars) and in general for content creation in gaming, film and AI. It comes as no surprise that there is significant interest in methods for automatic content generation~\cite{kar2019metasim,khalifa2020pcgrl} and in trying to bridge the performance gap of the models trained on synthetic data and deployed to the real world~\cite{ben2010theory,ganin2016domain}. 
Creating synthetic datasets comes with its own hurdles. While content, such as 3D CAD models that make up a scene are available on online asset stores, artists write complex procedural models that synthesize scenes by placing these assets in realistic layouts. This often requires browsing through massive amounts of real imagery to carefully tune a procedural model -- a time consuming task. For scenarios such as street scenes, creating synthetic scenes relevant for one city may require tuning a procedural model made for another city from scratch. In this paper, we propose an automatic method to carry out this task. 
%~\cite{kar2019metasim} hypothesized the presence of a \emph{content gap} and a \emph{style gap} between synthetic and real imagery. The content gap lies in the differences in distributions of objects and their relationships in scenes, whereas the style gap lies in the differences in low-level visual distributions. Bridging the \emph{style gap} has received a lot of attention~\cite{munit,unit} -- we aim to work towards bridging the \emph{content gap}. 

\begin{figure}[t!]
\centering
\includegraphics[width=0.85\linewidth, trim=240 240 160 50,clip]{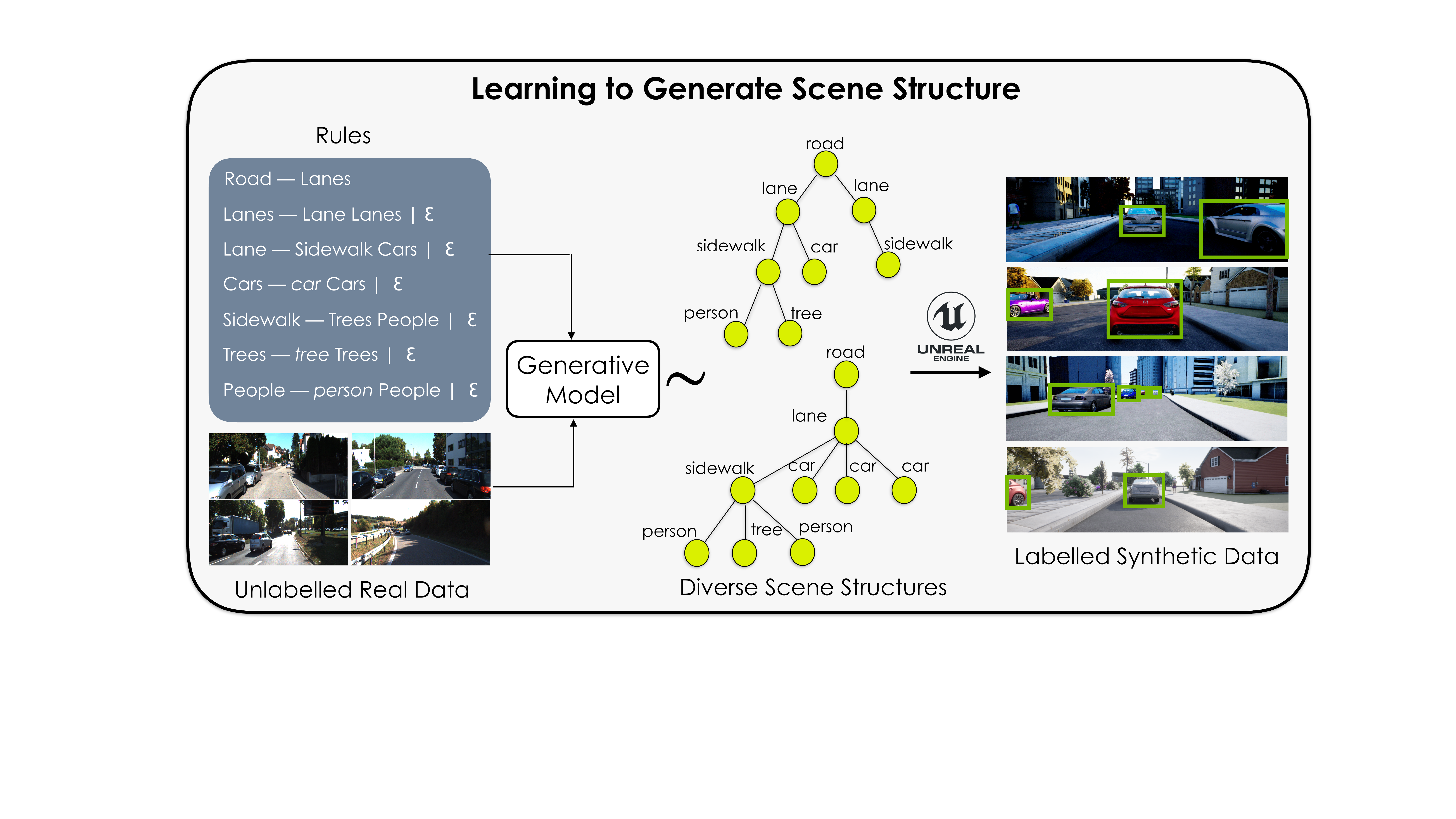}
\caption{We present a method that learns to generate synthetic scenes from real imagery in an unsupervised fashion. It does so by learning a generative model of scene structure, samples from which (with additional scene parameters) can be rendered to create synthetic images and labels.}
\end{figure}
%Since  have the potential to be massive in scale -- by 
%Recent work in synthetic content generation~\cite{kar2019metasim,DBLP:discrim} has allowed researchers to bypass expensive data collection and labelling. 
%Synthetic dataset generation requires generating discrete structural data such as 3D scene layouts. This consists of a \emph{scene structure} that defines the kinds of objects present in a scene and their quantity, and \emph{scene parameters} that defines how these objects are placed in the scene. Usually this is either manually crafted by artists, taken from real world examples~\cite{VirtualKITTI} or generated using carefully designed sampling procedures~\cite{sdr18}. In all of these situations, an expert is still required. This presents a major obstacle in applying synthetic content generation to previously unexplored domains.

%Tuning the parameters of these sampling procedures is already not an easy feat: for example, widths and curvatures of roads differ significantly in Manhattan or a village in Italy. Furthermore, structural data obtained from predefined sampling procedures are often much less complex than real examples. 
Recently, Meta-Sim~\cite{kar2019metasim} proposed %to alleviate the content gap by proposing 
to optimize scene parameters in a synthetically generated scene by exploiting the visual similarity of (rendered) generated synthetic data with real data. They represent scene structure and parameters in a \emph{scene graph}, and generate data by sampling a random scene structure (and parameters) from a given \emph{probabilistic grammar} of scenes, and then modifying the scene parameters using a learnt model. Since they only learn scene parameters, a sim-to-real gap in the scene structure remains. For example, one would likely find more cars, people and buildings in Manhattan over a quaint village in Italy. Other work on generative models of structural data such as graphs and grammar strings~\cite{Kusner:2017:GVA:3305381.3305582,dai2018syntax,DBLP:LGGAN,li2018learning} require large amounts of ground truth data for training to generate realistic samples. However, scene structures are extremely cumbersome to annotate and thus not available in most real datasets. 
%motivating our focus on learning them unsupervised.  

In this paper, we propose a procedural generative model of synthetic scenes that is learned unsupervised from real imagery. We generate \emph{scene graphs} object-by-object by learning to sample rule expansions from a given probabilistic scene grammar and generate scene parameters using~\cite{kar2019metasim}. %Due to the lack of annotated data, 
Learning without supervision here is a challenging problem due to the discrete nature of the scene structures we aim to generate and the presence of a non-differentiable renderer in the generative process.
To this end, we propose a feature space divergence to compare (rendered) generated scenes with real scenes, which can be computed per scene and is key to allowing credit assignment for training with reinforcement learning.

We evaluate our method on two synthetic datasets and a real driving dataset and find that our approach significantly reduces the distribution gap between scene structures in our generated and target data, improving over human priors on scene structure by learning to closely align with target structure distributions. On the real driving dataset, starting from minimal human priors, we show that we can almost exactly recover the structural distribution in the real target scenes (measured using GT annotations available for cars) -- an exciting result given that the model is trained without any labels. We show that an object detector trained on our generated data outperforms those trained on data generated with human priors or by ~\cite{kar2019metasim}, and show improvements in distribution similarity measures of our generated rendered images with real data.

\section{Related Work}
\label{related_work}

\subsection{Synthetic Content Creation}
Synthetic content creation has been receiving significant interest as a promising alternative to dataset collection and annotation. Various works have proposed generating synthetic data for tasks such as perception and planning in driving~\cite{RosCVPR16,VirtualKITTI,Richter16,Dosovitskiy17,sdr18,synscapes,alhaija2018augmented}, indoor scene perception~\cite{House3D,Zhang_2017_CVPR,mccormac2016scenenet,savva2019habitat,song2016ssc,handa2016understanding,armeni_iccv19}, game playing~\cite{openaigym,juliani2018unity}, robotic control~\cite{deepmindcontrolsuite2018} \cite{openaigym,MuJoCo,sadeghi2016cad2rl}, optical flow estimation~\cite{Sintel,shugrina2019creative}, home robotics~\cite{VirtualHome,THOR,VRKitchen} amongst many others, utilizing procedural modeling, existing simulators or human generated scenarios.

\textbf{Learnt Scene Generation} brings a data-driven nature to scene generation. ~\cite{craigyu2011furniture,wang2019planit} propose learning hierarchical spatial priors between furniture, that is integrated into a hand-crafted cost used to generate optimized indoor scene layouts.~\cite{qi2018human} similarly learn to synthesize indoor scenes using a probabilistic scene grammar and human-centric learning by leveraging affordances.~\cite{wang2019planit} learn to generate intermediate object relationship graphs and instantiate scenes conditioned on them.~\cite{Zhou_2019_ICCV} use a scene graph representation and learn adding objects into existing scenes.~\cite{Ritchie_2019_CVPR,li2019grains,wang2018deep} propose methods for learning deep priors from data for indoor scene synthesis.~\cite{eslami2016attend} introduce a generative model that sequentially adds objects into scenes, while ~\cite{Jyothi_2019_ICCV} propose a generative model for object layouts in 2D given a label set.~\cite{such2019generative} generate batches of data using a neural network that is used to train a task model, and learn by differentiating through the learning process of the task model.~\cite{kar2019metasim} propose learning to generate scenes by modifying the parameters of objects in scenes that are sampled from a probabilistic scene grammar. We argue that this ignores learning structural aspects of the scene, which we focus on in our work. Similar to~\cite{kar2019metasim,eslami2016attend}, and contrary to other works, we learn this in an unsupervised manner \ie given only target images as input.

\textbf{Learning with Simulators:} Methods in Approximate Bayesian Inference have looked into inferring the parameters of a simulator that generate a particular data point~\cite{mansinghka2013approximate,kulkarni2015picture}.~\cite{cranmer2019frontier} provide a great overview of advances in simulator based inference. Instead of running inference per scene~\cite{kulkarni2015picture,nsd}, we aim to generate new data that resembles a target distribution.~\cite{louppe2017adversarial} learn to optimize non-differentiable simulators using a variational upper bound of a GAN-like objective.~\cite{chebotar2018closing} learn to optimize simulator parameters for robotic control tasks by directly comparing trajectories between a real and a simulated robot.~\cite{ganin2018synthesizing,mellor2019unsupervised} train an agent to paint images using brush strokes in an adversarial setting with Reinforcement Learning. We learn to generate discrete scene structures constrained to a grammar, while optimizing a distribution matching objective (with Reinforcement Learning) instead of training adversarially. Compared to~\cite{mellor2019unsupervised}, we generate large and complex scenes, as opposed to images of single objects or faces.

\subsection{Graph Generation}
Generative models of graphs and trees ~\cite{li2018learning,DBLP:LGGAN,you2018graphrnn,liao2019efficient,chu2019neural,alvarez2016tree} generally produce graphs with richer structure with more flexibility over grammar based models, but often fail to produce syntactically correct graphs for cases with a defined syntax such as programs and scene graphs. \textbf{Grammar based methods} have been used for a variety of tasks such as program translation~\cite{TreeRNN}, conditional program generation~\cite{DBLP:journals/corr/YinN17,DBLP:StructVAE}, grammar induction~\cite{DBLP:GrammarInduction} and generative modelling on structures with syntax~\cite{Kusner:2017:GVA:3305381.3305582,dai2018syntax}, such as molecules. These methods, however, assume access to ground-truth graph structures for learning. We take inspiration from these methods, but show how to learn our model unsupervised \ie without any ground truth scene graph annotations.%\SF{weird without supervision on ground truth?}.
\begin{figure}[t!]
    \centering
    \includegraphics[width=0.85\linewidth, trim=0 430 100 0, clip]{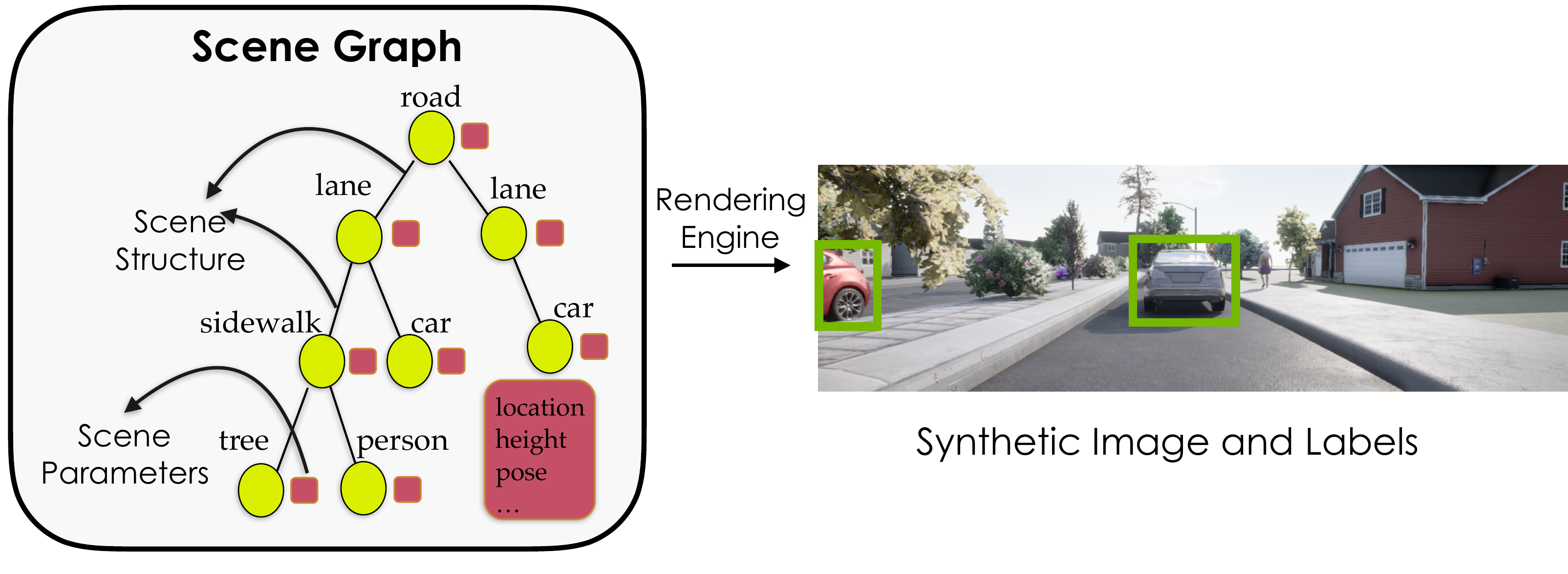}
    \caption{Example scene graph (structure and parameters) and depiction of its rendering}
    \label{fig:scene_graph}
\end{figure}

\section{Methodology}
\label{method}
We aim to learn a generative model of synthetic scenes. In particular, given a dataset of real imagery
$X_R$, the problem is to create synthetic data $D(\theta) = (X(\theta), Y(\theta))$ of images $X(\theta)$ and labels $Y(\theta)$ that is representative of $X_R$, where $\theta$ represents the parameters of the generative model. We exploit advances in graphics engines and rendering, by stipulating that the synthetic data $D$ is the output of creating an abstract scene representation and rendering it with a graphics engine. Rendering ensures that low level pixel information in $X(\theta)$ (and its corresponding annotation $Y(\theta)$) does not need to be modelled, which has been the focus of recent research in generative modeling of images~\cite{karras2018style,razavi2019generating}. Ensuring the semantic \emph{validity} of sampled scenes requires imposing constraints on their structure. Scene grammars use a set of rules to greatly reduce the space of scenes that can be sampled, making learning a more structured and tractable problem. For example, it could explicitly enforce that a car can only be on a road which then need not be implicitly learned, thus leading us to use probabilistic scene grammars. Meta-Sim~\cite{kar2019metasim} sampled \emph{scene graph} structures (see Fig.~\ref{fig:scene_graph}) from a prior imposed on a Probabilistic Context-Free Grammar (PCFG), which we call the \emph{structure prior}. They sampled parameters for every node in the scene graph from a \emph{parameter prior} and learned to predict new parameters for each node, keeping the structure intact. Their generated scenes therefore come from a structure prior (which is context-free) and the learnt parameter distribution, resulting in an untackled sim-to-real gap in the scene structures. In our work, we aim to alleviate this by learning a context-dependent structure distribution \emph{unsupervised} of synthetic scenes from images. %\SF{ a sentence with the main aim, ie learn ...}

We utilize \emph{scene graphs} as our abstract scene representation, that are rendered into a corresponding image with labels (Sec~\ref{ss:scene_repr}). Our generative model sequentially samples expansion rules from a given probabilistic scene grammar (Sec~\ref{ss:gen_model}) to generate a scene graph which is rendered. We train the model \emph{unsupervised} and with reinforcement learning, using a feature-matching based distribution divergence specifically designed to be amenable to our setting (Sec~\ref{ss:training}).

\begin{figure*}[t!]
\centering
\includegraphics[width=0.99\linewidth, trim=0 230 690 0,clip]{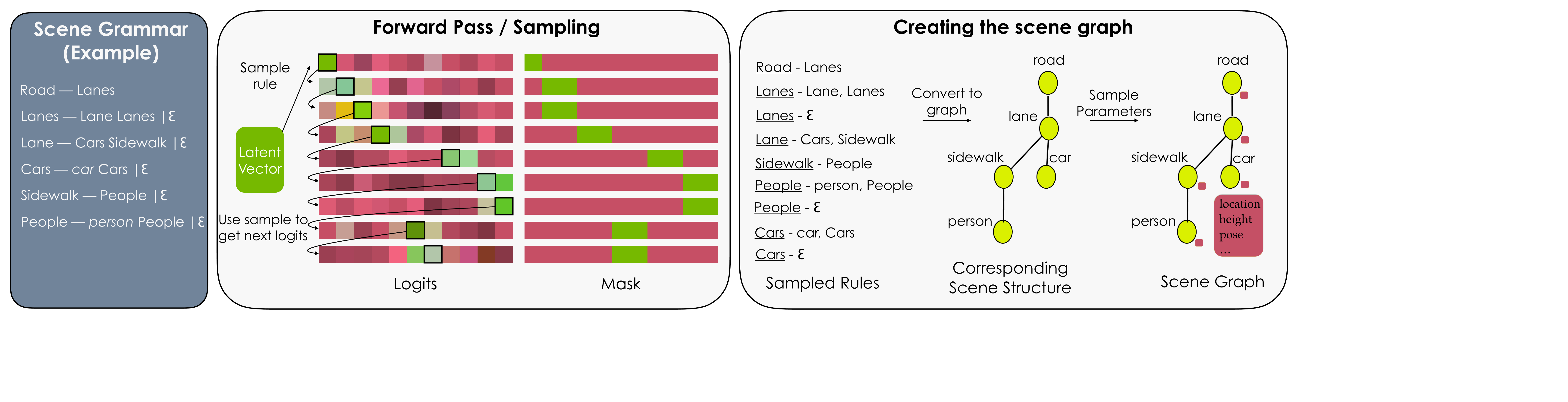}
\label{fig:fwd_pass}
\caption{Representation of our generative process for a scene graph. The logits and mask are of shape $T_{max} \times K$. Green represents a higher value and red is lower. At every time step, we autoregressively sample a rule and predict the logits for the next rule conditioned on the sample (capturing context dependencies). The figure on the right shows how sampled rules from the grammar are converted into a graph structure (only objects that are renderable are kept from the full grammar string). Parameters for every node can be sampled from a prior or optionally learnt with the method of~\cite{kar2019metasim}. A generated scene graph can be rendered as shown in Fig.~\ref{fig:scene_graph}.}
\end{figure*}

\subsection{Representing Synthetic Scenes}
\label{ss:scene_repr}
In Computer Graphics and Vision, \textbf{Scene Graphs} are commonly used to describe scenes in a concise hierarchical manner, where each node describes an object in the scene along with its parameters such as the 3D asset, pose etc. Parent-child relationships define the child's parameters relative to its parent, enabling easy scene editing and manipulation. Additionally, camera, lighting,
weather etc. are also encoded into the scene graph. Generating corresponding
pixels and annotations amounts to placing objects into the scene in a graphics engine
and rendering with the defined parameters (see Fig.~\ref{fig:scene_graph}).

%Ensuring the semantic \emph{validity} of scene graphs often requires imposing additional 
%constraints on their structure.~\cite{zhu2007stochastic,kar2019metasim} have proposed the use of
%\emph{probabilistic grammars} to sample \emph{diverse} and \emph{valid} scenes. Such a restriction greatly reduces
%the space of scenes that can be sampled, making learning a more structured and tractable problem. For example, a grammar might explicitly enforce that a car object can only be placed on a road, and not on a sidewalk -- resulting in the model not having to learn this relation implicitly. Because of these properties, we utilize scene graphs and probabilistic grammars in our generative model of synthetic scenes.

\textbf{Notation:} A context-free grammar $G$ is defined as a list of symbols (terminal and non-terminal) and expansion rules. Non-terminal symbols have at least one expansion rule into a new set of symbols. Sampling from a grammar involves expanding a start symbol till only non-terminal symbols remain. We denote the total number of expansion rules in a grammar $G$ as $K$. We define scene grammars and represent strings sampled from the grammar as scene graphs following~\cite{sdr18,kar2019metasim} (see Fig.~\ref{fig:fwd_pass}). For each scene graph, a structure $T$ is sampled from the grammar $G$ followed by sampling corresponding parameters $\alpha$ for every node in the graph. 

% To complete. Talk about the grammar, and define max number of grammar symbols $K$ etc. Talk about real and generated image distribution $p_I$, $q_I$.
% Talk about terminal and non-terminal symbols, the rendering function R
\subsection{Generative Model}
\label{ss:gen_model}
We take inspiration from previous work on learning generative models of graphs that are constrained by a grammar~\cite{Kusner:2017:GVA:3305381.3305582} for our architecture. Specifically, we map a latent vector $z$ to unnormalized probabilities over all possible grammar rules in an autoregressive manner, using a recurrent neural network till a maximum of $T_{max}$ steps. Deviating from~\cite{Kusner:2017:GVA:3305381.3305582}, we sample one rule $r_t$ at every time step and use it to predict the logits for the next rule $f_{t+1}$. This allows our model to capture context-dependent relationships easily, as opposed to the context-free nature of scene graphs in~\cite{kar2019metasim}. Given a list of at most $T_{max}$ sampled rules, the corresponding scene graph is generated by treating each rule expansion as a node expansion in the graph (see Fig.~\ref{fig:fwd_pass}).

\textbf{Sampling Correct Rules: } To ensure the validity of sampled rules in each time step $t$, we follow~\cite{Kusner:2017:GVA:3305381.3305582} and maintain a last-in-first-out (LIFO) stack of unexpanded non-terminal nodes. Nodes are popped from the stack, expanded according to the sampled rule-expansion, and the resulting new non-terminal nodes are pushed to the stack. When a non-terminal is popped, we create a mask $m_t$ of size $K$ which is 1 for valid rules from that non-terminal and 0 otherwise. Given the logits for the next expansion $f_t$, the probability of a rule $r_{t,k}$ is represented as,
\begin{align*}
p(r_t = k | f_t) = \frac{m_{t, k} exp(f_{t,k})}{\sum_{j=1}^{K} m_{t,j} exp(f_{t,j})}
\end{align*}
Sampling from this masked multinomial distribution ensures that only valid rules are sampled as $r_t$. 
Given the logits and sampled rules, $(f_t, r_t) \forall t \in {1 \ldots T_{max}}$, the
probability of the corresponding scene structure $T$ given $z$ is simply,
\begin{align*}
q_\theta(T|z) = \sum_{t=1}^{T_{max}} p(r_t|f_t)
\end{align*}
Putting it together, images are generated by sampling a scene structure $T \sim q_\theta(\cdot | z)$ from the model, followed by sampling parameters for every node in the scene $\alpha \sim q(\cdot | T)$ and rendering an image $v' = R(T, \alpha) \sim q_I$. For some $v' \sim q_I$, with parameters $\alpha$ and structure $T$, we assume\footnote{This equality does not hold in general for rendering, but it worked well in practice},
\begin{align*}
q_I(v'|z) = q(\alpha | T)q_\theta(T|z) 
\end{align*}

\subsection{Training}
\label{ss:training}
Training such a generative model is commonly done using \emph{variational inference}~\cite{kingma2013auto,rezende2014stochastic} 
or by optimizing a measure of \emph{distribution similarity}~\cite{goodfellow2014generative,li2015generative,li2017mmd,kar2019metasim}.
Variational Inference allows using reconstruction based objectives by introducing an approximate learnt posterior. Our attempts at using variational inference to train this model failed due to the
complexity coming from discrete sampling and having a renderer in the generative process. Moreover,
the recognition network here would amount to doing inverse graphics -- an extremely
challenging problem~\cite{kulkarni2015deep} in itself. Hence, we optimize a measure of distribution similarity of the generated and target data. We do not explore using a \emph{trained critic} due to the clear visual discrepancy between rendered and real images that a critic can exploit. Moreover, adversarial training is known to be notoriously difficult for discrete data. We note that recent work~\cite{ganin2018synthesizing,mellor2019unsupervised} has succeeded in adversarial training of a generative model of discrete brush strokes with reinforcement learning (RL), by carefully limiting the critic's capacity. We similarly employ RL to train our discrete generative model of scene graphs. While \emph{two sample tests}, such as MMD~\cite{gretton2012kernel} have been used in previous work to estimate and minimize the distance between two empirical distributions~\cite{li2015generative,dziugaite2015training,li2017mmd,kar2019metasim}, training with MMD and RL resulted in credit-assignment issues as it is a single score for the similarity of two full sets(batches) of data. Instead, our metric can be computed for every sample, which greatly helps training as shown empirically in Sec.~\ref{experiments}.

\textbf{Distribution Matching:}
We train the generative model to match the distribution of features of the real data 
in the latent space of some feature extractor $\varphi$. We define the real feature distribution $p_f$ s.t $F \sim p_f \iff F = \varphi(v)$ 
for some $v \sim p_I$. Similarly we define the generated feature distribution $q_f$ s.t $F \sim q_f \iff F = \varphi(v)$ 
for some $v \sim q_I$. We accomplish distribution matching by approximately computing $p_f, q_f$ from samples and minimizing the KL divergence from $p_f$ to 
$q_f$. Our training objective is 
\begin{align*}
    \min_\theta&\quad KL(q_f || p_f)\\
    \min_\theta&\quad \mathbb{E}_{F \sim q_f}[\log q_f(F) - \log p_f(F)]
\end{align*}
Using the feature distribution definition above, we have the equivalent objective
\begin{equation}
    \label{eq:dist_match}
    \min_\theta \mathbb{E}_{v \sim q_I} [\log q_f(\varphi(v)) - \log p_f(\varphi(v))]
\end{equation}

The true underlying feature distributions $q_f$ and $p_f$ are usually
intractable to compute. We use approximations $\tilde{q}_f(F)$ and $\tilde{p}_f(F)$, computed using kernel density estimation (KDE).
Let $V = \{v_1, \ldots, v_l\}$ and $B = \{v'_1, \ldots, v'_m\}$ be a batch of real and generated images.  KDE with $B,V$ to estimate $q_f, p_f$ yield
\begin{align*}
    \tilde{q}_f(F) = \frac{1}{m} \sum_{j = 1}^m K_{H}(F - \varphi(v_j'))\\
    \tilde{p}_f(F) = \frac{1}{l} \sum_{j = 1}^l K_{H}(F - \varphi(v_j))
\end{align*}
where $K_H$ is the standard multivariate normal kernel with bandwidth matrix $H$. We use $H = dI$ where $d$ is the dimensionality of the feature space. 

Our generative model involves making a discrete (non-differentiable) choice at each 
step, leading us to optimize our objective using reinforcement learning techniques\footnote{We did not explore sampling from a continuous relaxation of the discrete variable here}. 
Specifically, using the REINFORCE~\cite{williams1992simple} score function estimator along with a moving average baseline, we approximate the gradients of Eq.~\ref{eq:dist_match} as
\begin{equation}
    \nabla_\theta \mathcal{L} \approx \frac{1}{M} \sum_{j = 1}^m (\log\tilde{q}_f(\varphi(v_j')) - \log\tilde{p}_f(\varphi(v_j'))) \nabla_\theta \log q_I (v_j')
\end{equation}
where M is the batch size, $\tilde{q}_f(F)$ and $\tilde{p}_f(F)$ are density estimates defined above. 

Notice that the gradient above requires computing the marginal 
probability $q_I(v')$ of a generated image $v'$, instead of the conditional
$q_I(v'|z)$. Computing the \textbf{marginal probability of a generated image} requires an intractable marginalization over the latent 
variable $z$. To circumvent this, we use a fixed finite number of latent vectors from a set $Z$ sampled uniformly, enabling easy marginalization.
This translates to,
\begin{align*}
q_\theta(T) &= \frac{1}{|Z|} \sum_{z \in Z} q_\theta(T|z)\\
q_I(v') &= q(\alpha | T)q_\theta(T)
\end{align*}
We find that this still has enough modeling capacity, since there are only finitely many scene graphs of a maximum
length $T_{max}$ that can be sampled from the grammar. Empirically, we find using one latent vector to be enough in our
experiments. Essentially, stochasticity in the rule sampling makes up for lost stochasticity in the latent space. 

% \textbf{Discussion:}  
%We do not explore using a \emph{trained critic} due to the clear visual discrepancy between rendered and real images that the critic can exploit. Moreover, adversarial training is known to be notoriously difficult for discrete data. We note that recent work~\cite{ganin2018synthesizing,mellor2019unsupervised} has succeeded in training a generative model of discrete brush strokes with Reinforcement Learning by carefully limiting the trained critic's capacity. 
 %\SF{why not discuss this before you introduce your method?}. \AK{My reasoning was that this paragraph talks about optimizing MMD with RL being an issue. We introduce using RL only 1 paragraph ago, and therefore I felt it might be better to have it after.}

\textbf{Pretraining} is an essential step for our method. In every experiment, we define a simple handcrafted
prior on scene structure. For example, a simple prior could be to put one car on one road in a driving
scene. We pre-train the model by sampling strings (scene graphs) from the grammar prior, and training
the model to maximize the log-likelihood of these scene graphs. We provide specific details about the priors
used in Sec~\ref{experiments}.

\textbf{Feature Extraction} for distribution matching is a crucial step since
the features need to capture structural scene information such as the number of objects and their contextual spatial relationships for effective training.
We describe the feature extractor used and its training for each experiment in 
Sec~\ref{experiments}.

\textbf{Ensuring termination:} During training, sampling can result in incomplete strings generated with at most $T_\text{max}$ steps. Thus, we repeately sample a scene graph $T$ until its length is at most $T_\text{max}$. To ensure that we do not require too many attempts, we record the rejection rate $r_\text{reject}(F)$ of a sampled feature $F$ as the average failed sampling attempts when sampling the single scene graph used to generate $F$. We set a threshold $\epsilon$ on $r_\text{reject}(F)$ (representing the maximum allowable rejections) and weight $\lambda$ and add it to our original loss as,
\begin{align*}
\mathcal{L}' = \mathbb{E}_{F \sim q_F}[\log q_f(F) - \log p_f(F) + \lambda \mathbf{1}_{(\epsilon,\infty)}(r_\text{reject}(F))]
\end{align*}
We found that $\lambda = 10^{-2}$ and $\epsilon = 1$ worked well for all of our experiments.

% \SF{Since below are other or failed options, maybe put this in a discussion? right now its bolded and looks as part of method}

\section{Experiments}
\label{experiments}

We show two controlled experiments, on the MNIST dataset~\cite{lecun1998mnist} (Sec.~\ref{ss:exp_mnist}) and on synthetic aerial imagery~\cite{kar2019metasim} (Sec.~\ref{ss:exp_aerial}), where we showcase the ability of our model to learn synthetic structure distributions \textbf{unsupervised}. Finally, we show an experiment on generating 3D driving scenes (Sec.~\ref{ss:exp_kitti}), mimicking structure distributions on the KITTI~\cite{kitti} driving dataset and showing the performance of an object detector trained on our generated data. The renderers used in each experiment are adapted from~\cite{kar2019metasim}.
For each experiment, we first dicuss the corresponding scene grammar. Then, we discuss the feature extractor and its training. Finally, we describe the structure prior used to pre-train the model, the target data, and show results on learning to mimic structures in the target data without any access to ground-truth structures. Additionally, we show comparisons with learning with MMD~\cite{gretton2012kernel} (Sec.~\ref{ss:exp_mnist}) and show how our model can learn to generate context-dependent scene graphs from the grammar (Sec.~\ref{ss:exp_aerial}).

\subsection{Multi MNIST}
\label{ss:exp_mnist}
We first evaluate our approach on a toy example of learning to generate scenes with multiple digits. The grammar defining the scene structure is:
\begin{align*}
    \text{Scene} \to bg \ \text{Digits}, \quad \text{Digits} \to \text{Digit} \ \text{Digits} \ | \ \epsilon, \quad
    \text{Digit} \to 0 \ | \  1 \ | \ 2 \ | \ \cdots \ | \ 9 
\end{align*}
Sampled digits are placed onto a black canvas of size $256\times256$.

\textbf{Feature Extraction Network:}
We train a network to determine the binary presence of a digit class in the scene. We use a Resnet~\cite{DBLP:journals/corr/HeZRS15} made up of three residual blocks each containing two $3\times3$ convolutional layers to produce an image embedding and three fully connected layers from the image embedding to make the prediction. We use the Resnet embeddings as our image features. We train the network on synthetic data generated by our simple prior for both structure and continuous parameters. Training is done with a simple binary cross-entropy criterion for each class. The exact prior and target data used is explained below.
% \subsubsection{Prior }
% We form two target sets of images by placing up to 4 digits next to each other horizontally on the canvas with even spacing. The first set (Figure 2) uses the digits from MNIST while the second (Figure 3) uses font based digits. We impose a categorical distribution on the digit class and number of digits in each scene. Both sets use the same distribution. The exact distribution can be seen in Fig X. The prior \AK{fill here} and can be seen in Fig. X.

\begin{figure}[t!]
  \centering
  \begin{minipage}{0.49\textwidth}
    \centering
    \includegraphics[width=0.4\linewidth, trim=0 30 0 30]{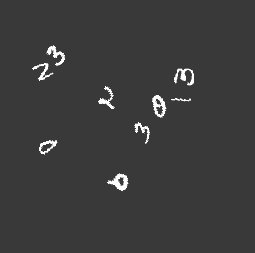}
    \includegraphics[width=0.4\linewidth, trim=0 30 0 30]{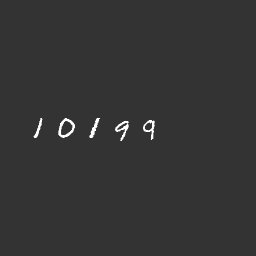}
    \caption{Prior (Left) and Validation (Right) example for MultiMNIST experiments}
    \label{fig:prior-samples-mnist}
  \end{minipage}
  \begin{minipage}{0.49\textwidth}
    \centering
    \includegraphics[width=0.4\linewidth, trim=0 30 0 30]{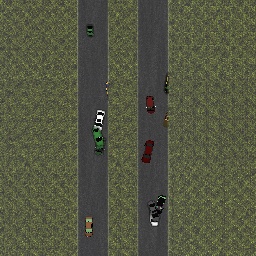}
    \includegraphics[width=0.4\linewidth, trim=0 30 0 30]{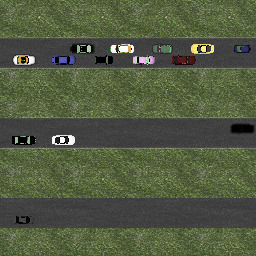}
    \caption{Prior (Left) and Validation (Right) example for Aerial 2D experiments}
    \label{fig:prior-samples-aerial}
  \end{minipage}
  \begin{minipage}{0.49\textwidth}
    \centering
    \includegraphics[width=0.49\linewidth, trim=0 40 0 0]{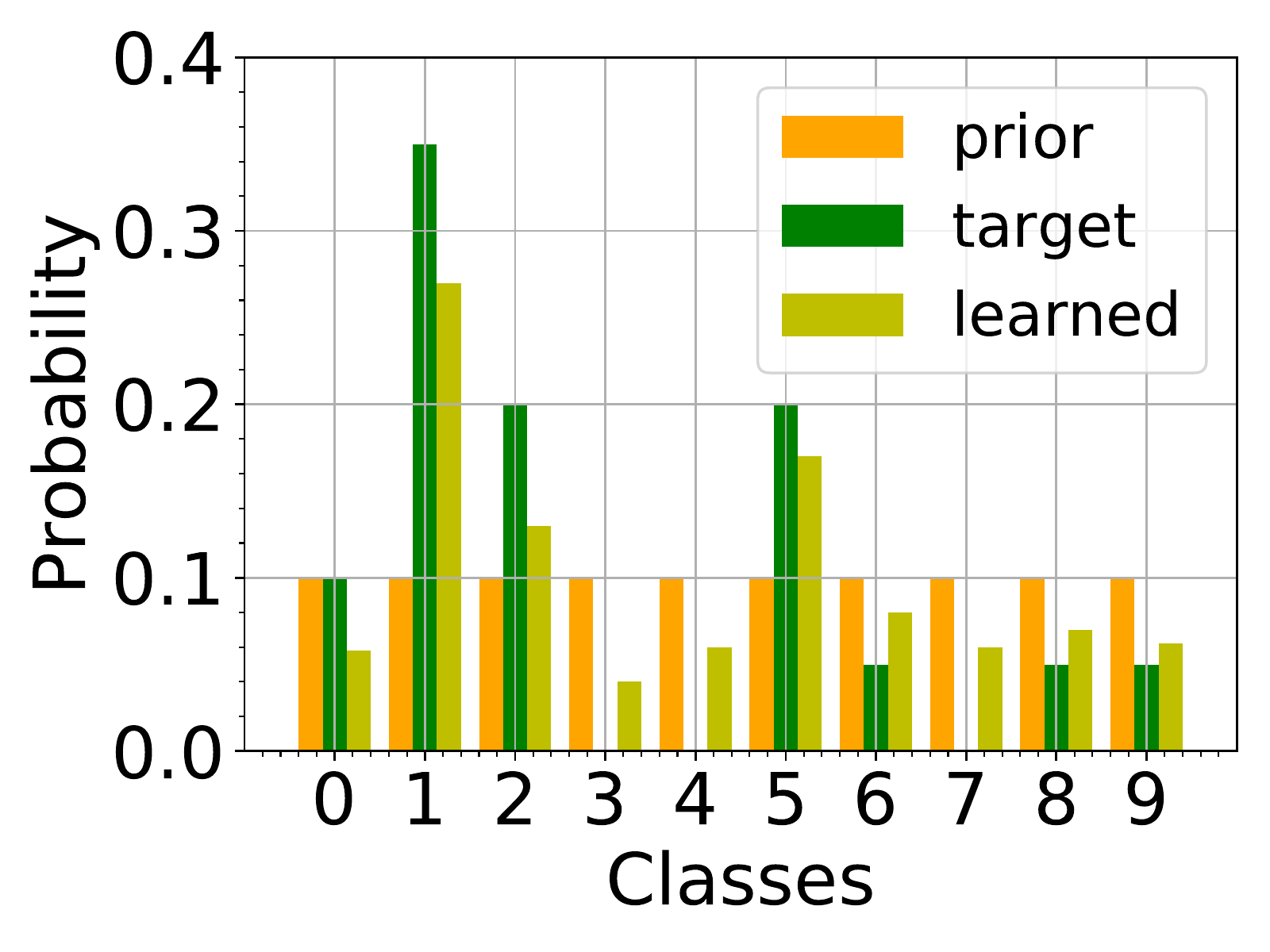}
    \includegraphics[width=0.49\linewidth, trim=0 40 0 0]{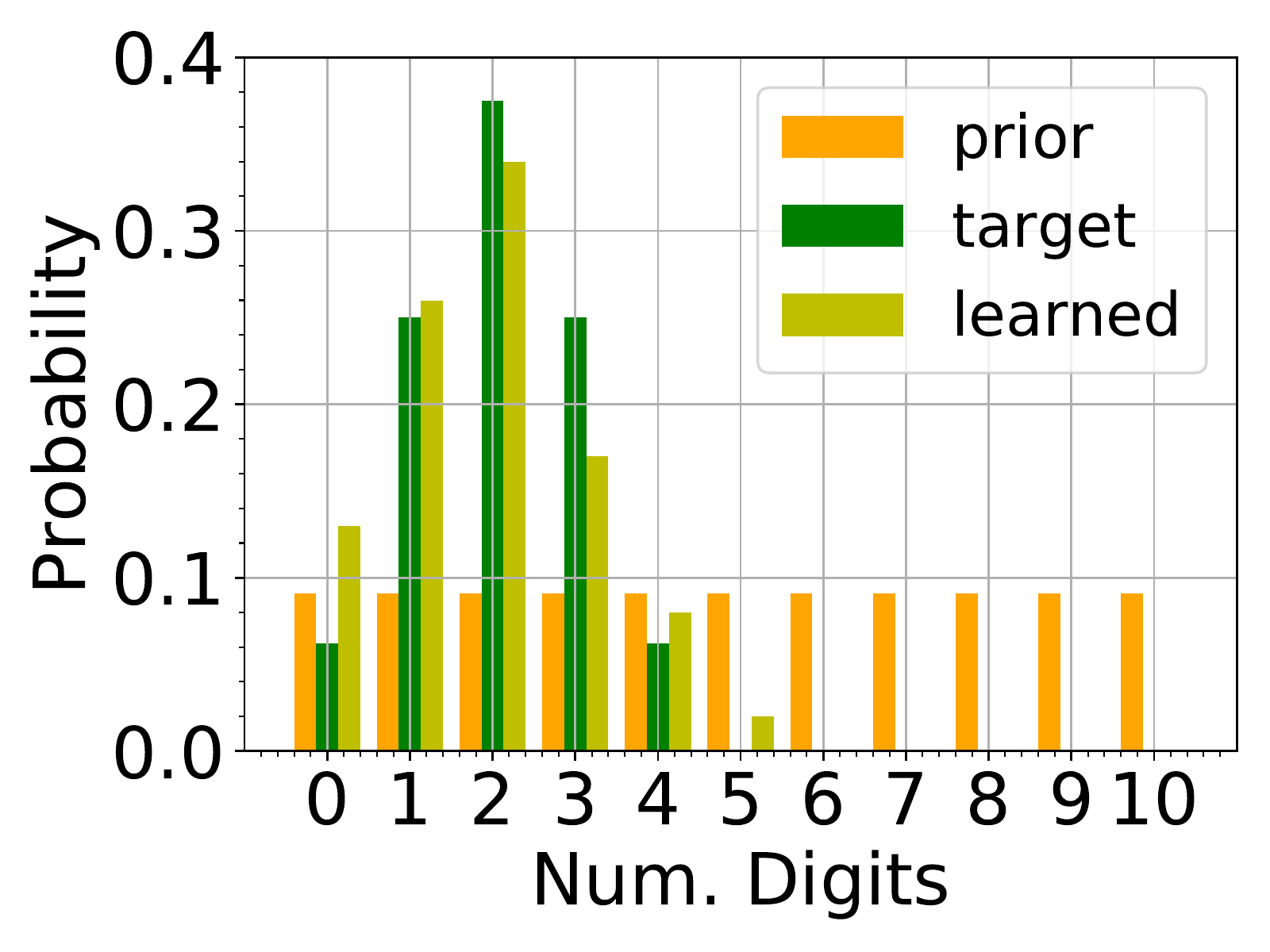}
    \caption{Distributions of classes and number of digits, in the prior, learned and target scene structures}
    \label{fig:mnist-uni}
  \end{minipage}
  \begin{minipage}{0.49\textwidth}
    \centering
    \includegraphics[width=0.49\linewidth, trim=0 40 0 0]{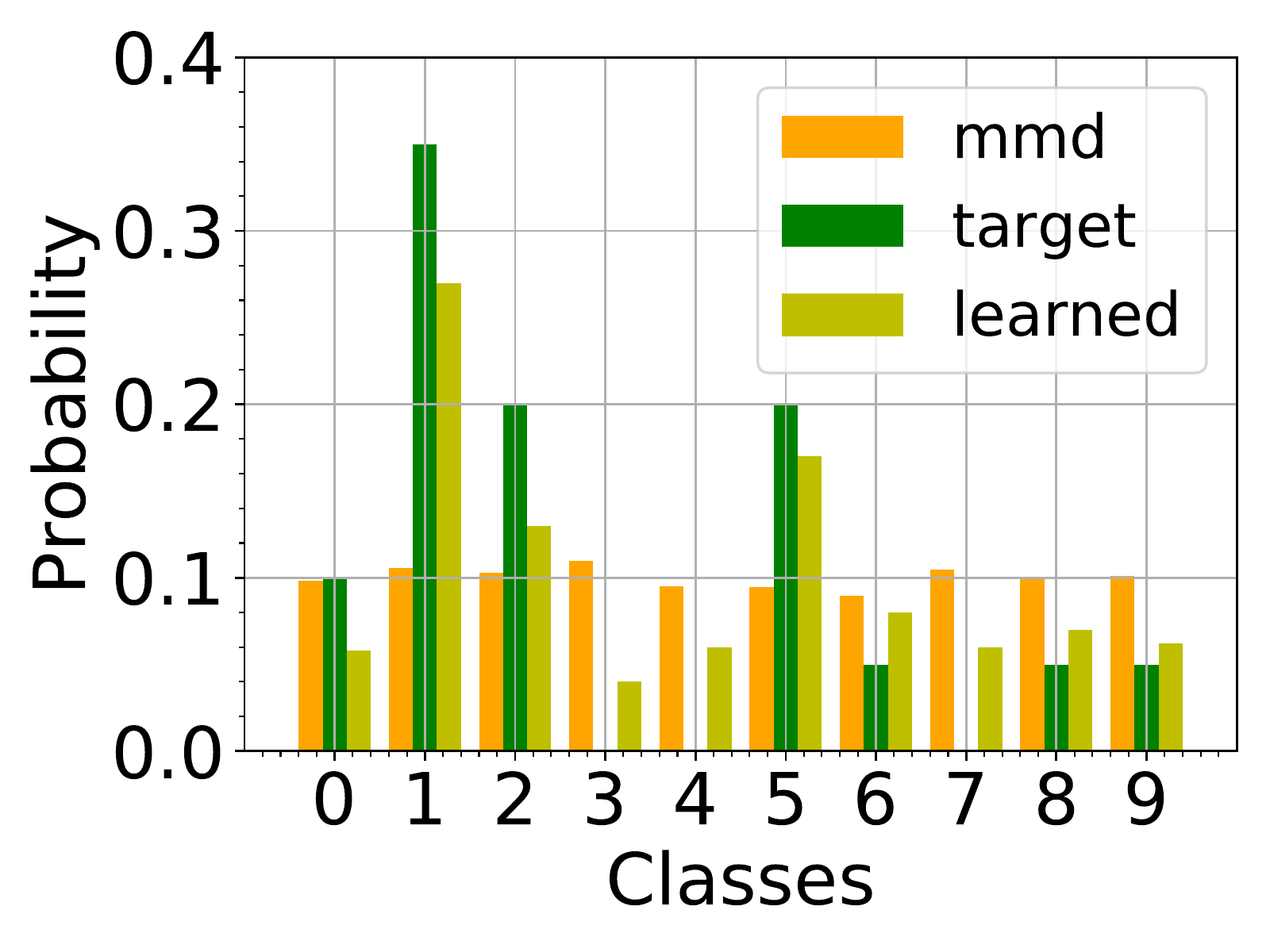}
    \includegraphics[width=0.49\linewidth, trim=0 40 0 0]{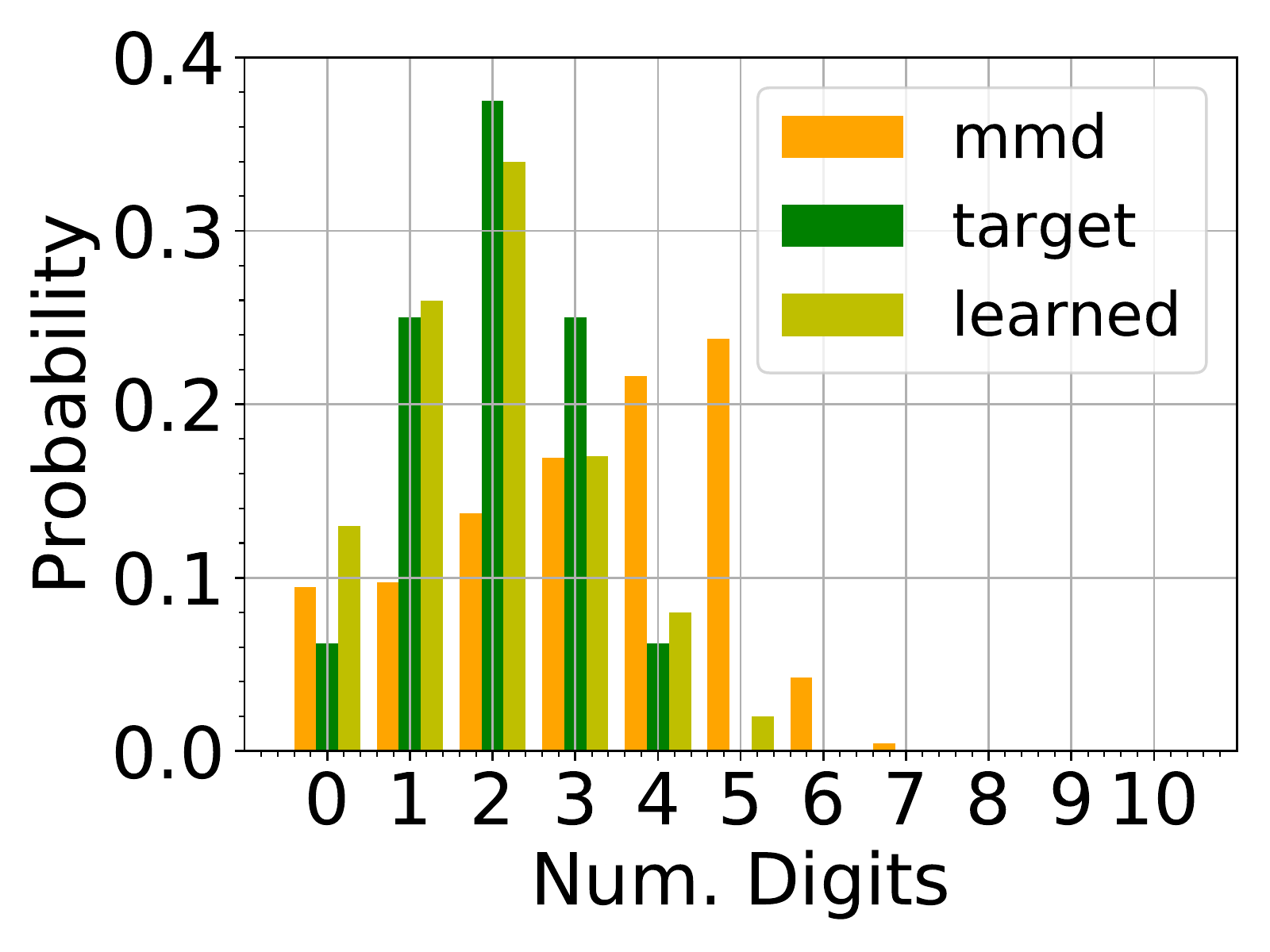}
    \caption{Distributions of classes and number of digits, comparing learning with MMD, ours and the target}
    \label{fig:mnist-mmd-compare}
  \end{minipage}    
\end{figure}

\textbf{Prior and Target Data:}
We sample the number of digits in the scene $n_d$ uniformly from $0$ to $10$, and sample $n_d$ digits uniformly to place on the scene. The digits are placed(parameters) uniformly on the canvas. The target data has digits upright in a straight line in the middle of the canvas. Fig.~\ref{fig:prior-samples-mnist} shows example prior samples, and target data. We show we can learn scene structures with a gap remaining in the parameters by using the parameter prior during training.

We attempt \textbf{learning a random distribution of number of digits} with random classes in the scene. Fig.~\ref{fig:mnist-uni} shows the prior, target and learnt distribution of the number of digits and their class distribution. We see that our model can faithfully approximate the target, even while learning it unsupervised. We also \textbf{train with MMD~\cite{gretton2012kernel}}, computed using two batches of real and generated images and used as the reward for every generated scene. Fig.~\ref{fig:mnist-mmd-compare} shows that using MMD results in the model learning a smoothed approximation of the target distribution, which comes from the lack of credit assignment in the score, that we get with our objective.

% \begin{figure}[h]
  % \centering
  % \begin{minipage}{0.49\textwidth}
  %   \centering
  %   \includegraphics[width=0.49\linewidth, trim=0 40 0 10]{mnist-classes-unimodal.pdf}
  %   \includegraphics[width=0.49\linewidth, trim=0 40 0 10]{mnist-num-digits-unimodal.pdf}
  %   \caption{Distributions of classes and number of digits, in the prior, learned and target scene structures}
  %   \label{fig:mnist-uni}
  % \end{minipage}
  % \begin{minipage}{0.49\textwidth}
  %   \centering
  %   \includegraphics[width=0.49\linewidth, trim=0 30 0 10]{mnist-mmd-compare-classes.pdf}
  %   \includegraphics[width=0.49\linewidth, trim=0 30 0 10]{mnist-mmd-compare.pdf}
  %   \caption{Distributions of classes and number of digits, comparing learning with MMD, ours and the target}
  %   \label{fig:mnist-mmd-compare}
  % \end{minipage}    
% \end{figure}

\subsection{Aerial 2D}
\label{ss:exp_aerial}
Next, we evaluate our approach on a harder synthetic scenario of aerial views of driving scenes. The grammar and the corresponding rendered scenes offer additional complexity to test the model. The grammar here is as follows:  
\begin{align*}
    &\text{Scene} \to \text{Roads}, \quad &\text{Roads} \to \text{Road} \ \text{Roads} \ | \ \epsilon\\
    &\text{Road} \to \text{Cars}, \quad &\text{Cars} \to car \ \text{Cars} \ | \ \epsilon
\end{align*}
\textbf{Feature Extraction Network:}
We use the same Resnet~\cite{DBLP:journals/corr/HeZRS15} architecture from the MNIST experiment with the FC layers outputting the number of cars, roads, houses and trees in the scene as 1-hot labels. We train by minimizing the cross entropies these labels, trained on samples generated from the prior.

\begin{figure*}[t!]
  \centering
  \hfill
  \includegraphics[width=0.32\textwidth, trim=0 50 0 30]{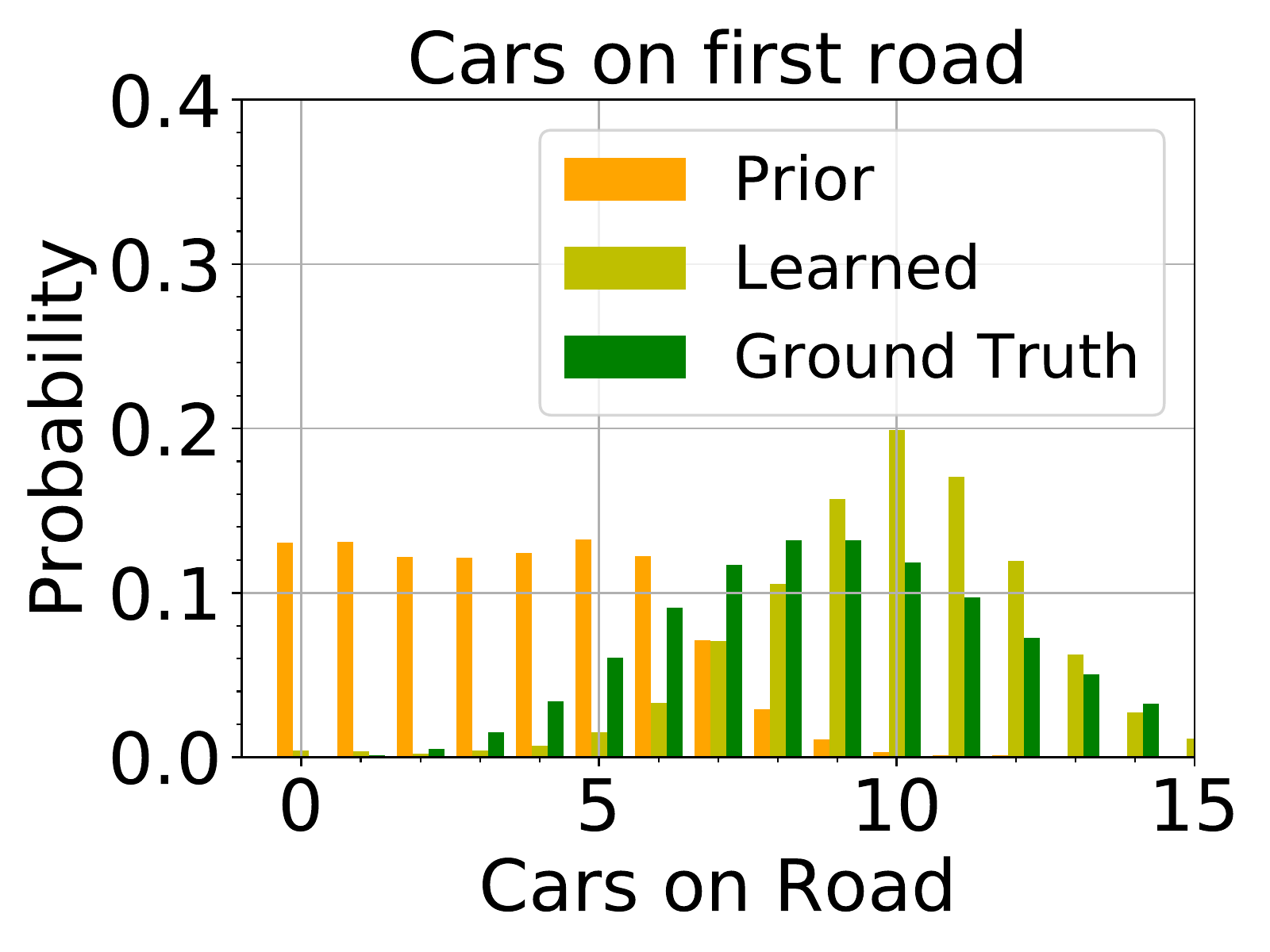}
  % \hfill
  % \includegraphics[width=0.32\textwidth]{KITTI_HISTS/suburban_3D.pdf}
  \hfill
  \includegraphics[width=0.32\textwidth, trim=0 50 0 30]{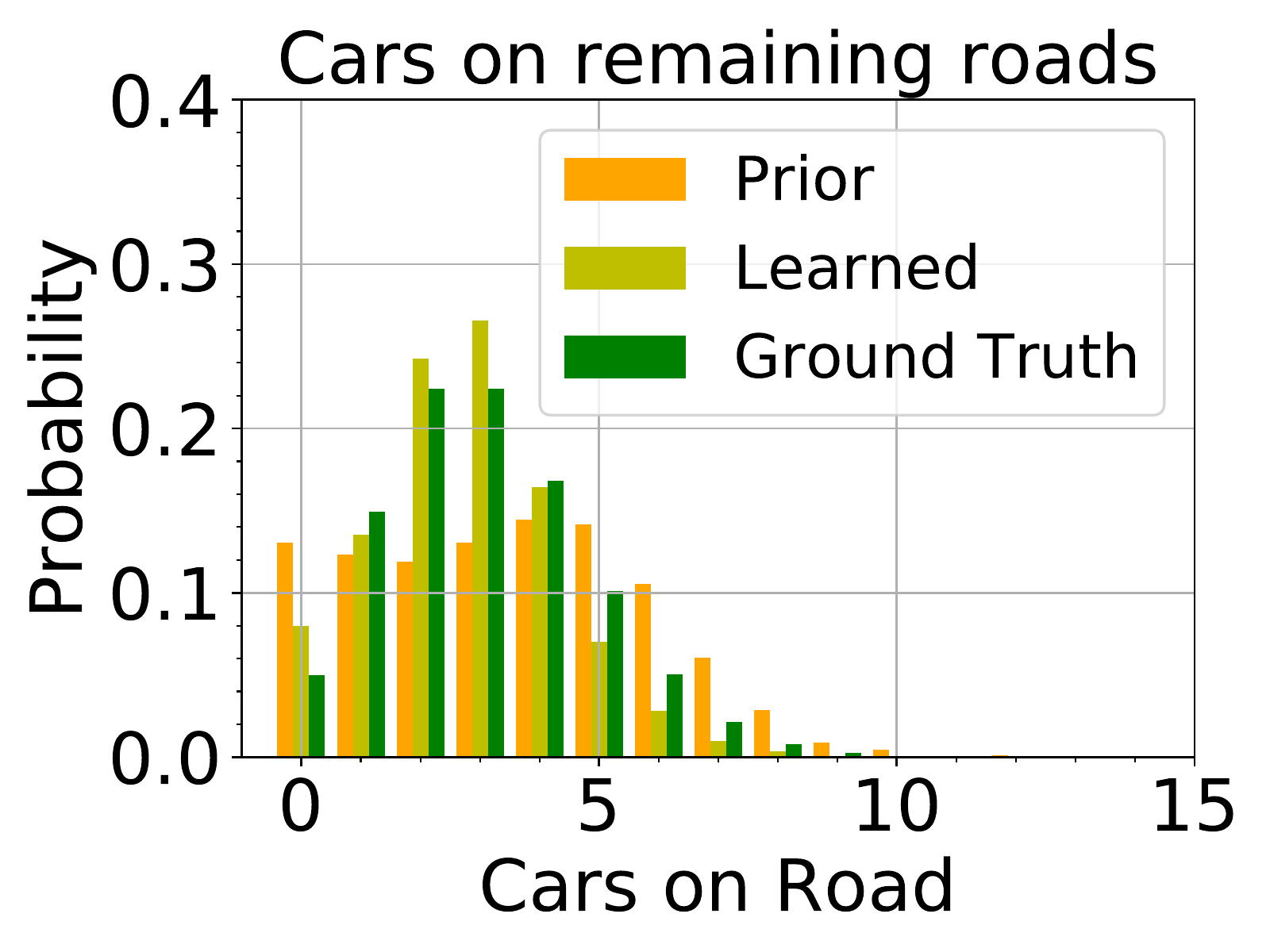}
  \hfill
  \hfill
  \caption{\#cars distribution learned in the Aerial 2D experiment. We can learn \emph{context dependent} relationships, placing different number of cars on different roads}
  \label{fig:aerial}
\end{figure*}

\textbf{Prior:}
We sample the number of roads $n_r \in [0, 4]$ uniformly. On each road, we sample $c \in [0, 8]$ cars uniformly. Roads are placed sequentially by sampling a random distance $d$ and placing the road $d$ pixels in front of the previous one. Cars are placed on the road with uniform random position and rotation (Fig.~\ref{fig:prior-samples-aerial}).

\textbf{Learning context-dependent relationships:}
For the target dataset, we sample the number of roads  $n_r \in [0, 4]$ with probabilities (0.05, 0.15, 0.4, 0.4). On the first road we sample $n_1 \sim \text{Poisson}(9)$ cars and $n_i \sim \text{Poisson}(3)$ cars for each of the remaining roads. All cars are placed well spaced on their respective road. Unlike the Multi-MNIST experiment, these structures cannot be modelled by a Probabilistic-CFG, and thus by ~\cite{Kusner:2017:GVA:3305381.3305582,kar2019metasim}. We see that our model can learn this context-dependent distribution faithfully as well in Fig.~\ref{fig:aerial}.

\subsection{3D Driving Scenes}
\label{ss:exp_kitti}
We experiment on the KITTI ~\cite{kitti} dataset, which was captured with a camera
mounted on top of a car driving around the city of Karlsruhe, Germany. The dataset contains a wide variety of road scenes, ranging from urban traffic scenarios to highways and more rural neighborhoods. We utilize the same grammar and renderer used for road scenes in~\cite{kar2019metasim}. Our model, although trained unsupervised, can learn to get closer to the underlying structure distribution, improve measures of image generation, and the performance of a downstream task model.

\textbf{Prior and Training:}
Following SDR~\cite{sdr18}, we define three different priors to capture three different modes in the KITTI dataset. They are the 'Rural', 'Suburban' and 'Urban' scenarios, as defined in~\cite{sdr18}. We train three different versions of our model, one for each of the structural priors, and sample from each of them uniformly. We use the scene parameter prior and learnt scene parameter model from~\cite{kar2019metasim} to produce parameters for our generated scene structures to get the final scene graphs, which are rendered and used for our distribution matching. 

\textbf{Feature Extraction Network:}
We use the pool-3 layer of an Inception-V3 network, pre-trained on the ImageNet~\cite{deng2009imagenet} dataset as our feature extractor. Interestingly, we found this to work as well as using features from Mask-RCNN~\cite{he2017mask} trained on driving scenes.

\begin{figure*}[t!]
  \centering
  \includegraphics[width=0.32\textwidth]{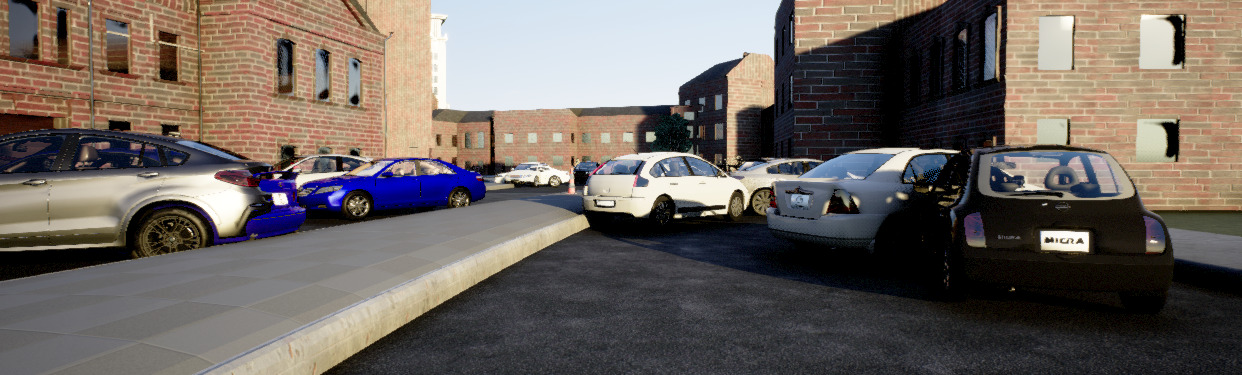}
  \includegraphics[width=0.32\textwidth]{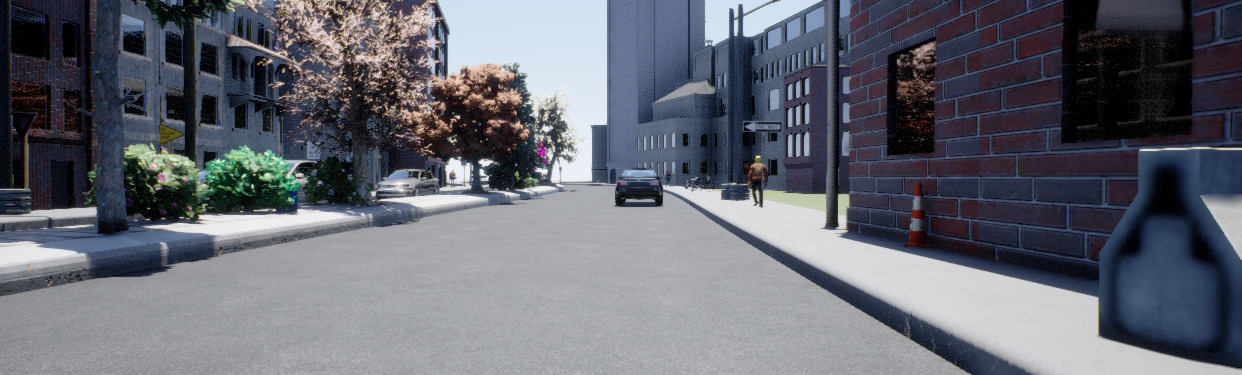}
  \includegraphics[width=0.32\textwidth]{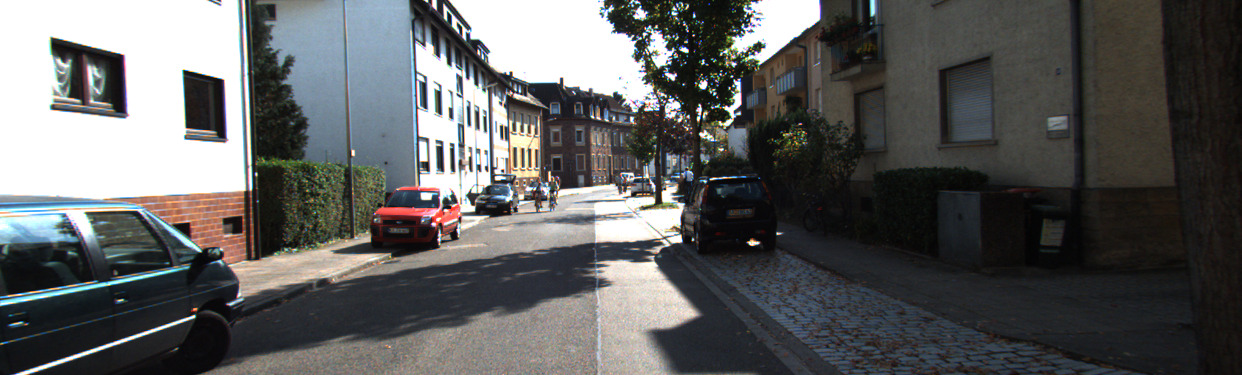}
  \includegraphics[width=0.32\textwidth]{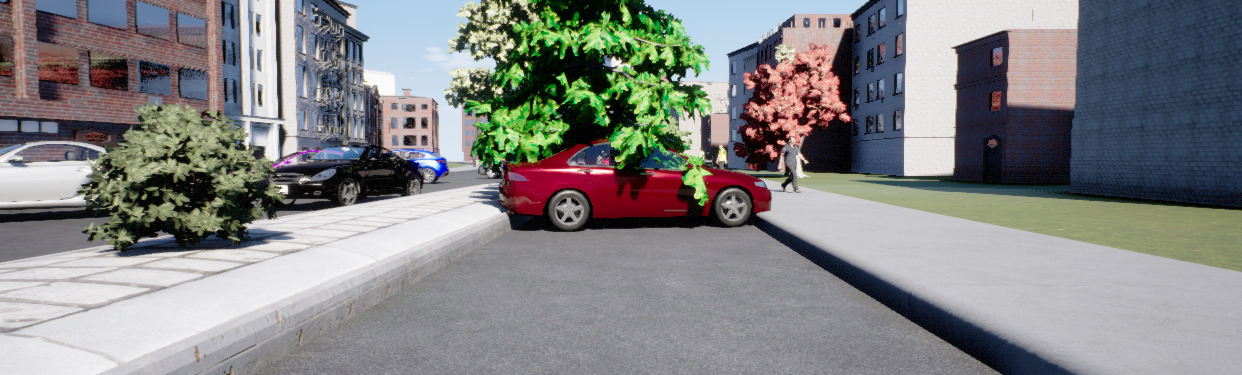}
  \includegraphics[width=0.32\textwidth]{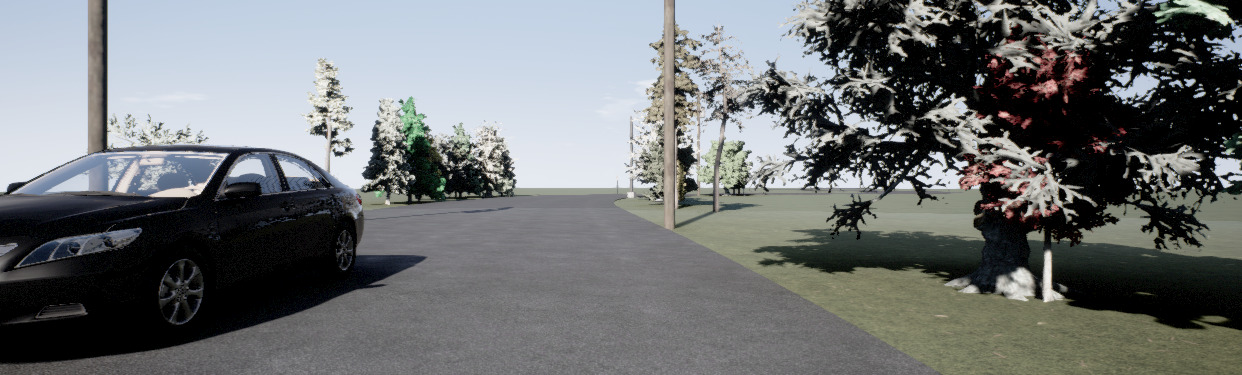}
  \includegraphics[width=0.32\textwidth]{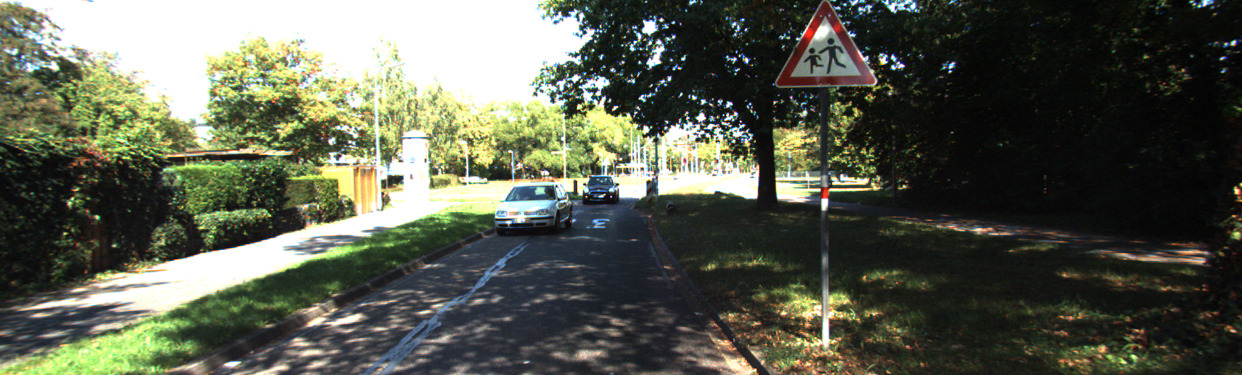}
  \includegraphics[width=0.32\textwidth]{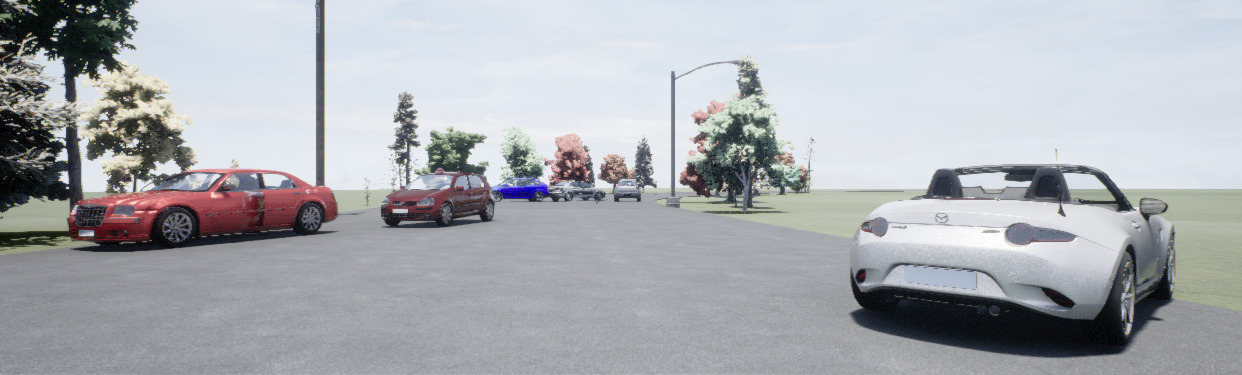}
  \includegraphics[width=0.32\textwidth]{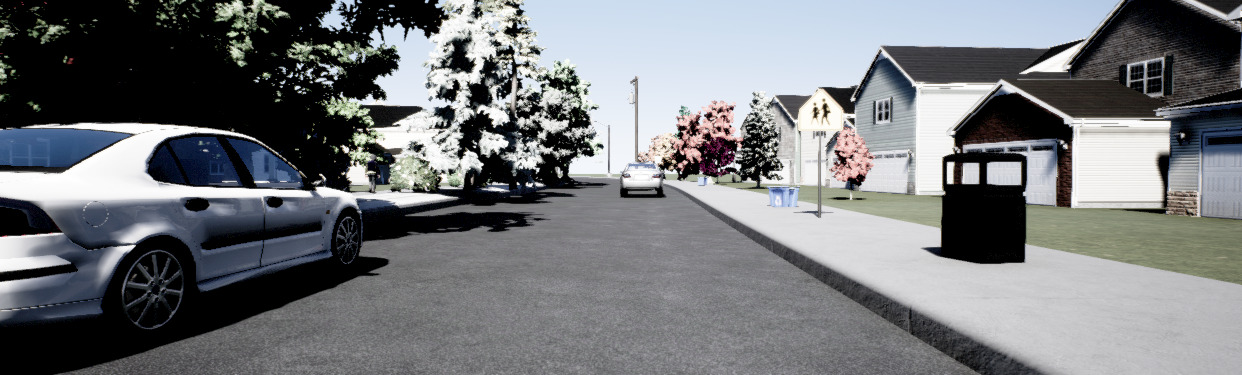}
  \includegraphics[width=0.32\textwidth]{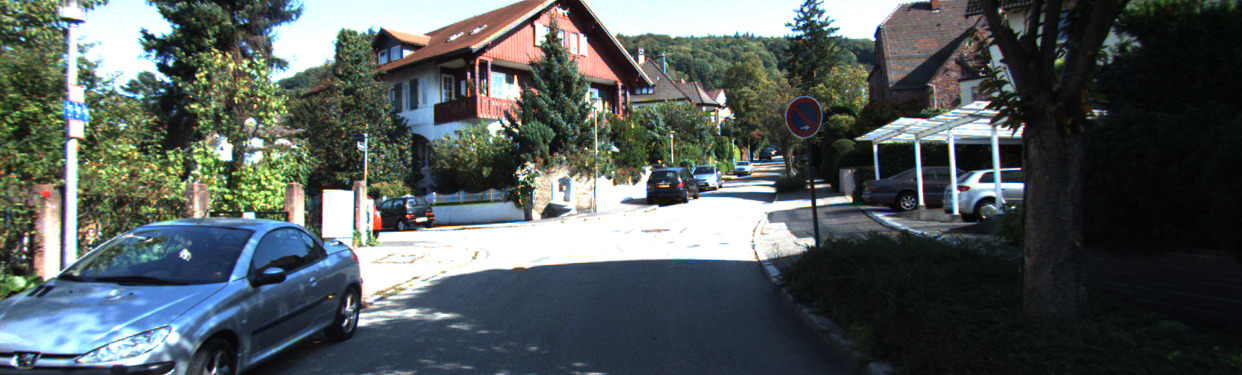}
  \includegraphics[width=0.32\textwidth, trim=0 30 0 0]{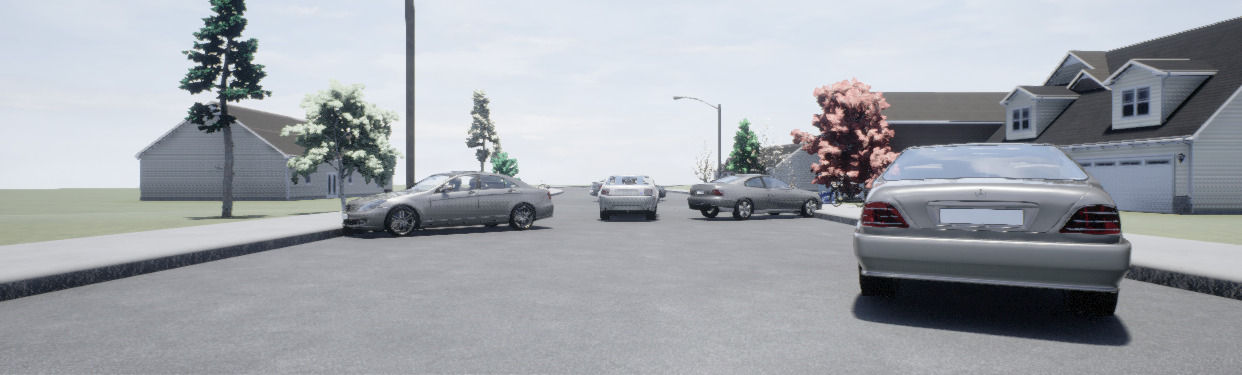}
  \includegraphics[width=0.32\textwidth, trim=0 30 0 0]{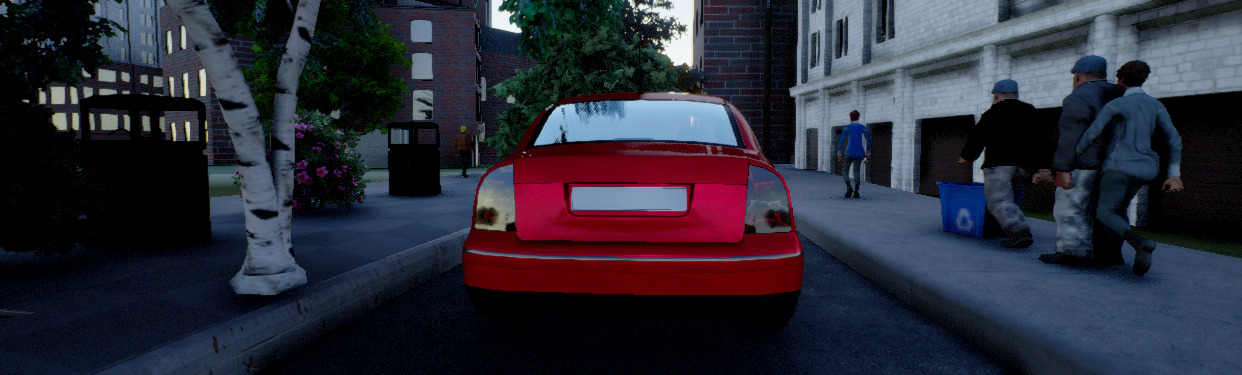}
  \includegraphics[width=0.32\textwidth, trim=0 30 0 0]{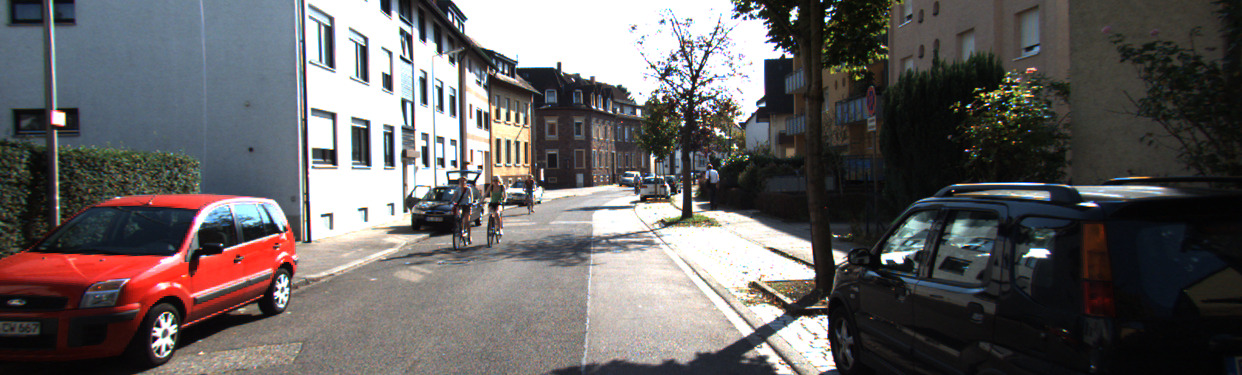}
  \caption{Generated images (good prior expt.). (Left) Using both the structure and parameter prior, (Middle) Using our learnt structure and parameters from~\cite{kar2019metasim}, (Right) Real KITTI samples. Our model (middle), although unsupervised, adds diverse scene elements like vegetation, pedestrians, signs etc. to better resemble the real dataset.}
  \label{fig:kitti_images}
\end{figure*}

\begin{figure*}[t!]
  \centering
  \hfill
  \includegraphics[width=0.32\textwidth, trim=0 50 0 30]{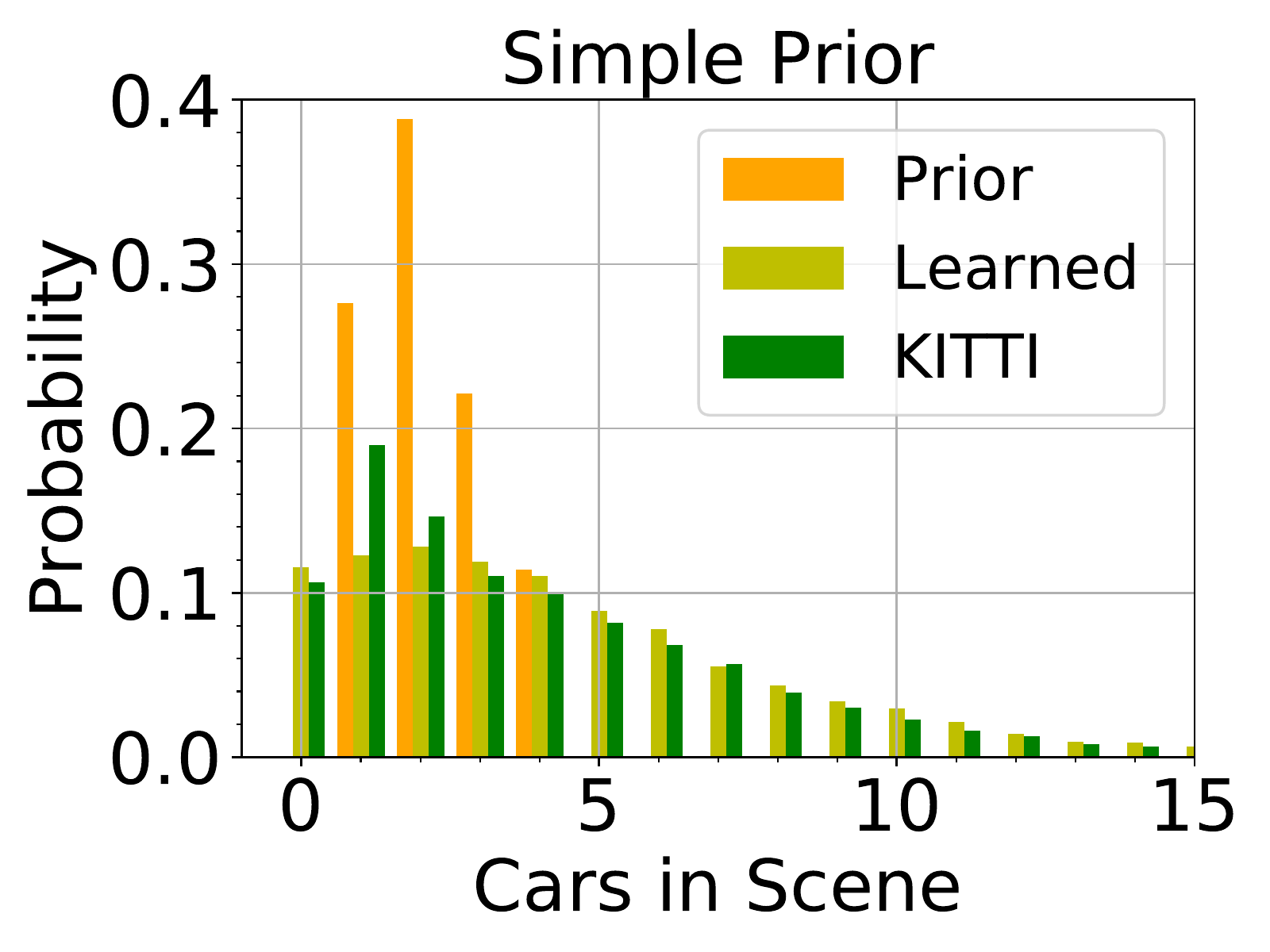}
  \hfill
  \includegraphics[width=0.32\textwidth, trim=0 50 0 30]{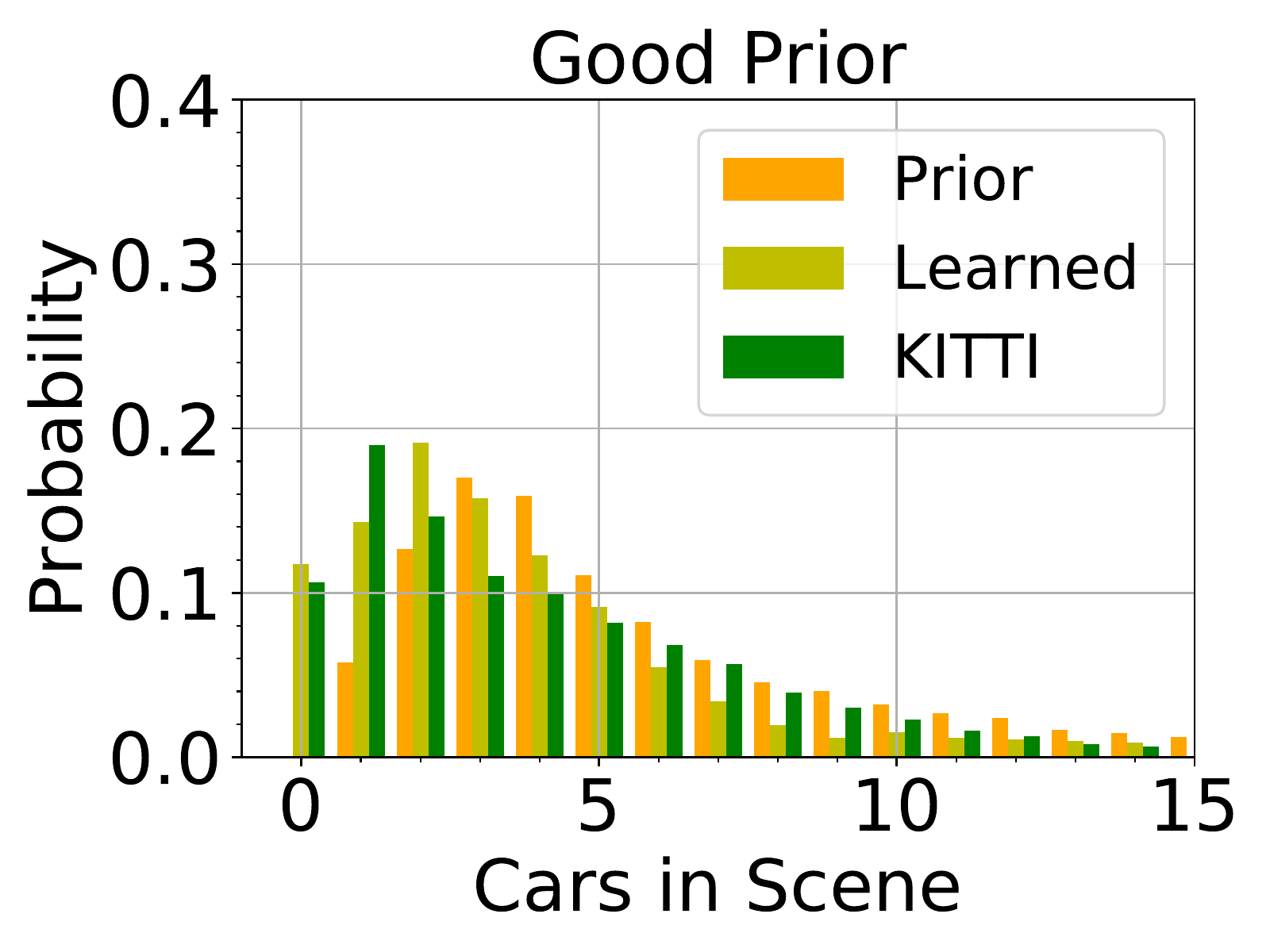}
  \hfill
  \hfill
  \caption{\#cars/scene learned from a simple prior (left) and good prior (right) on KITTI}
  \label{fig:kitti}
\end{figure*}

\begin{figure*}[t!]
  \centering
  \includegraphics[width=0.32\textwidth]{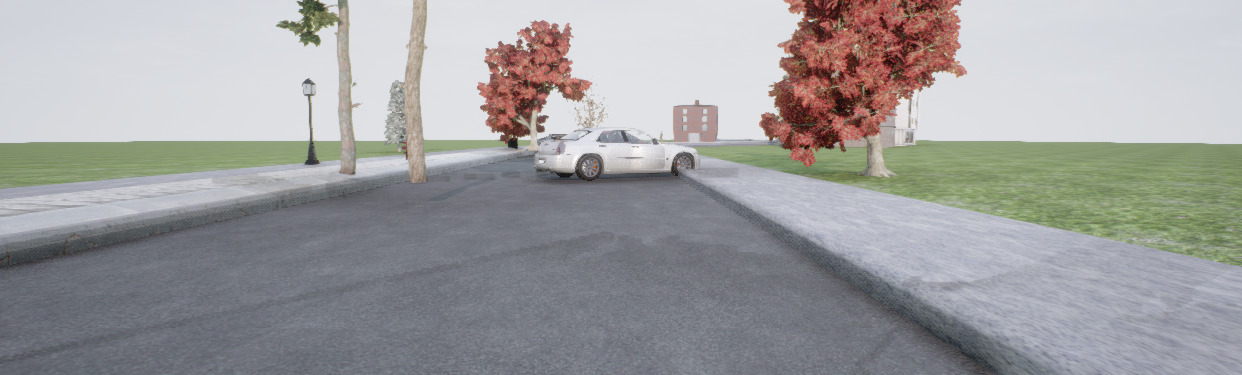}
  \includegraphics[width=0.32\textwidth]{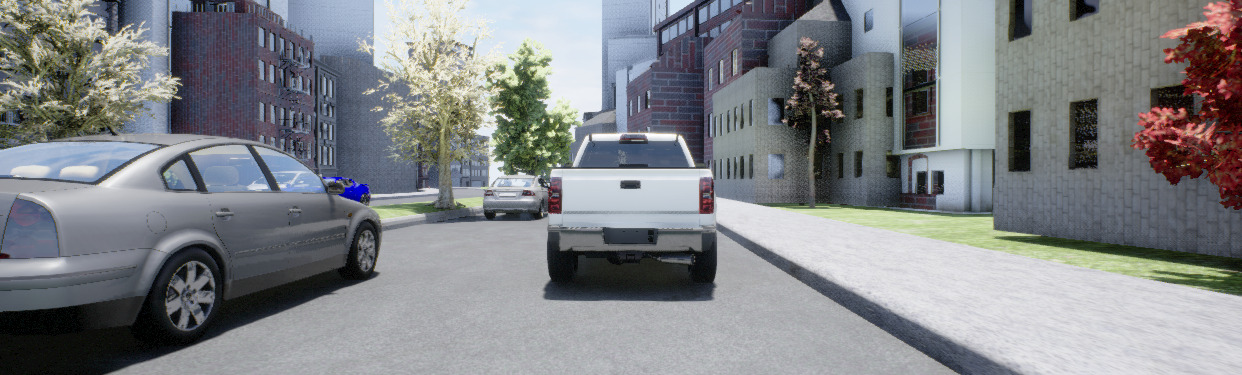}
  \includegraphics[width=0.32\textwidth]{Selection/kitti/004551.jpg}
  \includegraphics[width=0.32\textwidth]{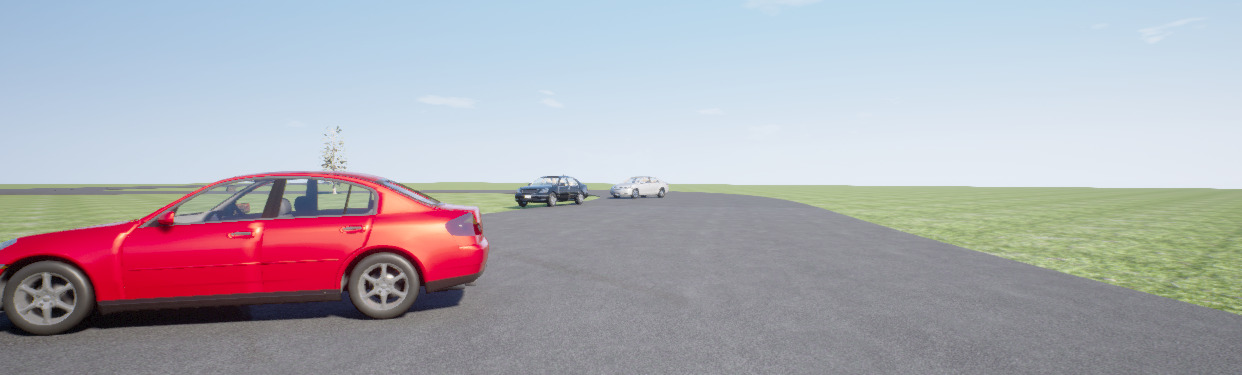}
  \includegraphics[width=0.32\textwidth]{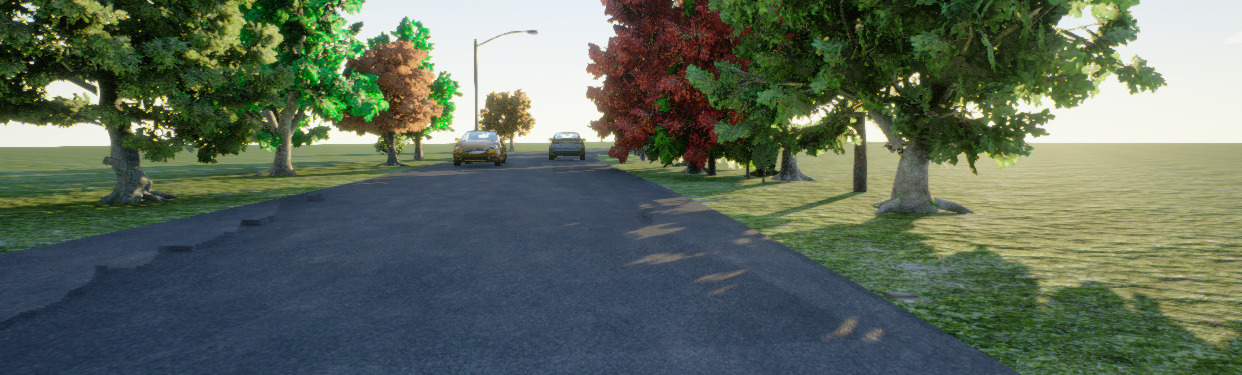}
  \includegraphics[width=0.32\textwidth]{Selection/kitti/rural.jpg}
  \includegraphics[width=0.32\textwidth]{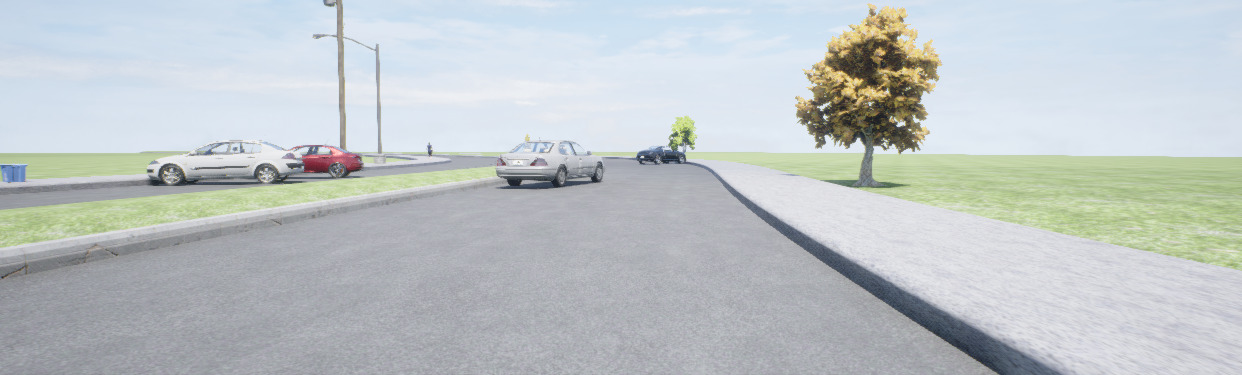}
  \includegraphics[width=0.32\textwidth]{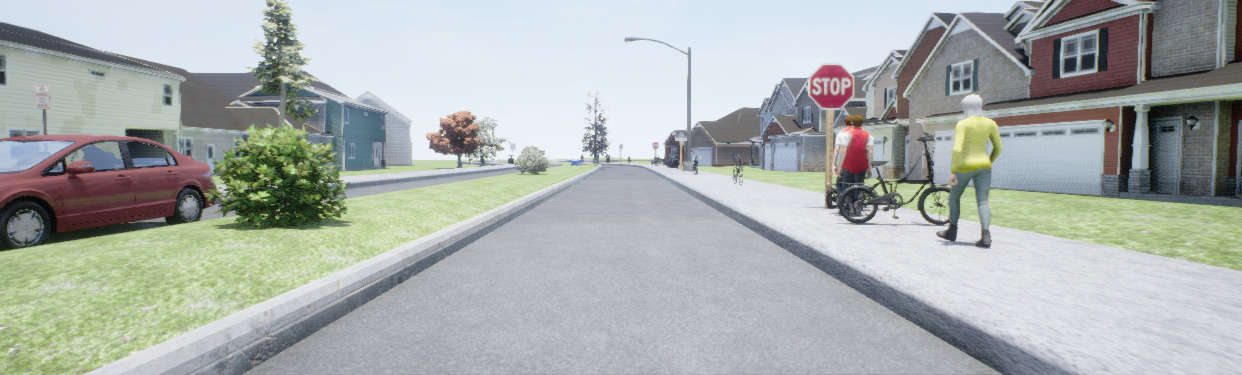}
  \includegraphics[width=0.32\textwidth]{Selection/kitti/suburban.jpg}
  \includegraphics[width=0.32\textwidth, trim=0 30 0 0]{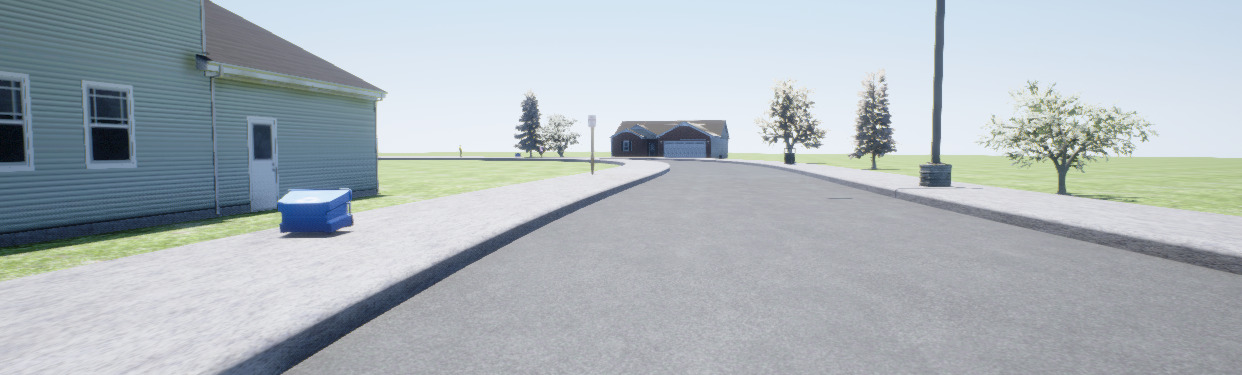}
  \includegraphics[width=0.32\textwidth, trim=0 30 0 0]{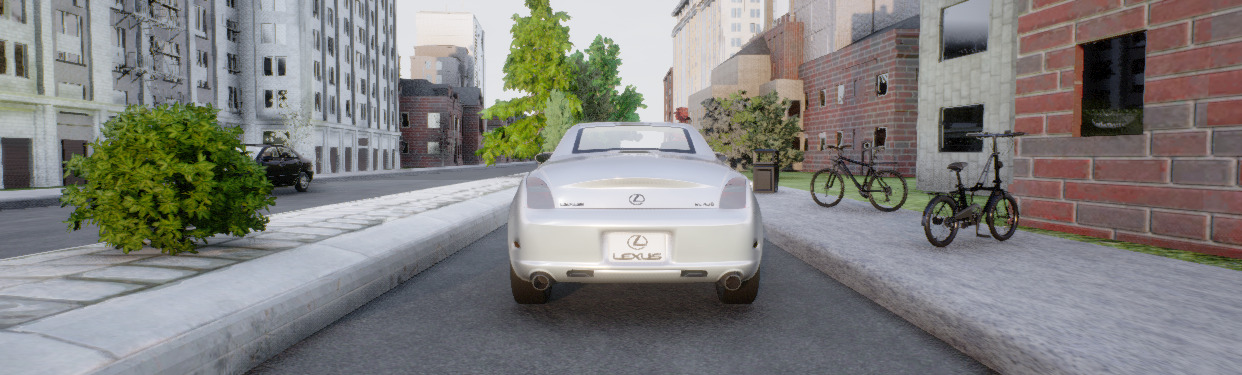}
  \includegraphics[width=0.32\textwidth, trim=0 30 0 0]{Selection/kitti/city_1.jpg}
  \caption{Generated images (simple prior expt.). (Left) Using both the structure and parameter prior, (Middle) Using our learnt structure and parameters from~\cite{kar2019metasim}, (Right) Real samples from KITTI. Our model, although trained unsupervised, learns to add an appropriate frequency and diversity of scene elements to resemble the real data, even when trained from a very weak prior.}
  \label{fig:kitti_bad_prior}
\end{figure*}

\textbf{Distribution similarity metrics:} In generative modeling of images, the Frechet Inception Distance~\cite{heusel2017gans}, and the Kernel Inception Distance~\cite{binkowski2018demystifying} have been used to measure progress. We report FID and KID in Tab.~\ref{tab:kitti},~\ref{tab:kitti_bad_prior} between our generated synthetic dataset and the KITTI-train set. We do so by generating 10K synthetic samples and using the full KITTI-train set, computed using the pool-3 features of an Inception-v3 network. Fig.~\ref{fig:kitti} (left) shows the distribution of the number of cars generated by the prior, learnt model and in the KITTI dataset (since we have GT for cars). We do not have ground truth for which KITTI scenes could be classified into rural/suburban/urban, so we compare against the global distribution of the whole dataset. We notice that the model bridges the gap between this particular distribution well after training. 

\textbf{Task Performance:} We report average precision for detection at 0.5 IoU \ie AP@0.5 (following~\cite{kar2019metasim}) of an object detector trained to convergence on our synthetic data and tested on the KITTI validation set. We use the detection head from Mask-RCNN \cite{he2017mask} with a Resnet-50-FPN backbone initialized with pre-trained ImageNet weights as our object detector. The task network in each result row of Tab.~\ref{tab:kitti} is finetuned from the snapshot of the previous row.~\cite{kar2019metasim} show results with adding Image-to-Image translation to the generated images to reduce the \emph{appearance gap} and results with training on a small amount of real data. We omit those experiments here and refer the reader to their paper for a sketch of expected results in these settings. Training this model directly on the full KITTI training set obtains AP@0.5 of $81.52(\text{easy})$, $83.58(\text{medium})$ and $84.48(\text{hard})$, denoting a large sim-to-real performance gap left to bridge.

% \begin{table}[h]
  % \centering
  % \begin{tabular}{c|c|c|c|c|c|c|c}
  %   \textbf{Method} & \textbf{Structure} & \textbf{Parameters} & \textbf{Easy} & \textbf{Medium} & \textbf{Hard} & \textbf{KID}~\cite{binkowski2018demystifying} & \textbf{FID}~\cite{heusel2017gans}\\
  %   \hline \hline
  %   Prob. Grammar & Prior & Prior & 63.7 & 63.7 & 62.2 & 0.066 & 106.6\\
  %   Meta-Sim*~\cite{kar2019metasim} & Prior & Learnt & 66.5 & 66.3 & 65.8 & 0.072 & 111.6\\
  %   Ours & Learnt & Learnt & \textbf{67.0} & \textbf{67.0} & \textbf{66.2} & \textbf{0.054} & \textbf{99.7}\\
  %   \hline
  % \end{tabular}
  % \caption{AP@0.5 on the KITTI-val set, and distribution similarity metrics between generated synthetic data and the KITTI-train set. Learnt parameters are used from~\cite{kar2019metasim}. *Results from ~\cite{kar2019metasim} are our reproduced numbers, and we show learning the structure additionally helps close the distribution gap and improves downstream task performance. See Fig.~\ref{fig:kitti_images} for qualitative examples}
  % \label{tab:kitti}
% \end{table}

\textbf{Using a simple prior:} The priors on the structure in the previous experiments were taken from~\cite{sdr18}. These priors already involved some human intervention, which we aim to minimize. Therefore, we repeat the experiments above with a very simple and quick to create prior on the scene structure, where a few instances of each kind of object (car, house etc.) is placed in the scene (see Fig.~\ref{fig:kitti_bad_prior} (Left)).~\cite{kar2019metasim} requires a decently crafted structure prior to train the parameter network. Thus, we use the prior parameters while training our structure generator in this experiment (showing the robustness of training with randomized prior parameters), and learn the parameter network later (Tab.~\ref{tab:kitti_bad_prior}). Fig.~\ref{fig:kitti} (right) shows that the method learned the distribution of the number of cars well (unsupervised), even when initialized from a bad prior. Notice that the FID/KID of the learnt model from the simple prior in Tab.~\ref{tab:kitti_bad_prior} is comparable to that trained from a tuned prior in Tab.~\ref{tab:kitti}, which we believe is an exciting result.

\begin{table}[t!]
  \centering
    \begin{tabular}{c|c|c|c|c|c|c|c}
    \textbf{Method} & \textbf{Structure} & \textbf{Parameters} & \textbf{Easy} & \textbf{Medium} & \textbf{Hard} & \textbf{KID}~\cite{binkowski2018demystifying} & \textbf{FID}~\cite{heusel2017gans}\\
    \hline \hline
    Prob. Grammar & Prior & Prior & 63.7 & 63.7 & 62.2 & 0.066 & 106.6\\
    Meta-Sim*~\cite{kar2019metasim} & Prior & Learnt & 66.5 & 66.3 & 65.8 & 0.072 & 111.6\\
    Ours & Learnt & Learnt & \textbf{67.0} & \textbf{67.0} & \textbf{66.2} & \textbf{0.054} & \textbf{99.7}\\
    \hline
  \end{tabular}
  \caption{AP@0.5 on KITTI-val and distribution similarity metrics between generated synthetic data and KITTI-train. Learnt parameters are used from~\cite{kar2019metasim}. *Results from ~\cite{kar2019metasim} are our reproduced numbers, and we show learning the structure additionally helps close the distribution gap and improves downstream task performance.}
  \label{tab:kitti}

  \begin{tabular}{c|c|c|c|c|c|c|c}
    \textbf{Method} & \textbf{Structure} & \textbf{Parameters} & \textbf{Easy} & \textbf{Medium} & \textbf{Hard} & \textbf{KID}~\cite{binkowski2018demystifying} & \textbf{FID}~\cite{heusel2017gans}\\
    \hline \hline
    Prob. Grammar & Prior* & Prior & 61.3 & 59.8 & 58.0 & 0.101 & 130.3\\
    Ours & Learnt & Prior & 63.2 & 62.5 & 61.2 & \textbf{0.059} & \textbf{100.0}\\
    Ours & Learnt & Learnt & \textbf{65.2} & \textbf{64.7} & \textbf{63.4} & 0.060 & 101.7\\
    \hline
  \end{tabular}
  \caption{Repeat of experiments in Tab.~\ref{tab:kitti} with a *\emph{simple} prior on the scene structure. Parameters are learnt using~\cite{kar2019metasim}. We observe a significant boost in both task performance and distribution similarity metrics, by learning the structure and parameters.}
  \label{tab:kitti_bad_prior}
\end{table}

\textbf{Discussion:} We noticed that our method worked better when initialized with more spread out priors than more localized priors (Tab.~\ref{tab:kitti},~\ref{tab:kitti_bad_prior}, Fig.~\ref{fig:kitti}) We hypothesize this is due to the distribution matching metric we use being the the reverse-KL divergence between the generated and real data  (feature) distributions, which is mode-seeking instead of being mode-covering. Therefore, an initialization with a narrow distribution around one of the modes has low incentive to move away from it, hampering learning. Even then, we see a significant improvement in performance when starting from a peaky prior as shown in Tab.~\ref{tab:kitti_bad_prior}. We also note the importance of pre-training the task network. Rows in Tab.~\ref{tab:kitti} and Tab.~\ref{tab:kitti_bad_prior} were finetuned from the checkpoint of the previous row. The first row (Prob. Grammar) is a form of Domain Randomization~\cite{drjosh17,sdr18}, which has been shown to be crucial for sim-to-real adaptation. Our method, in essence, reduces the randomization in the generated scenes (by learning to generate scenes similar to the target data), and we observe that progressively training the task network with our (more specialized) generated data improves its performance.~\cite{akkaya2019solving,wang2019poet} show the opposite behavior, where increasing randomization (or environment difficulty) through task training results in improved performance. A detailed analysis of this phenomemon is beyond the current scope and is left for future work.
\section{Conclusion}
We introduced an approach to unsupervised learning of a generative model of synthetic scene structures by optimizing for visual similarity to the real data. Inferring scene structures is known to be notoriously hard even when annotations are provided. Our method is able to perform the generative side of it without any ground truth information. Experiments on two toy and one real dataset showcase the ability of our model to learn a plausible posterior over scene structures, significantly improving over  manually designed priors. 
Our current method needs to optimize for both the scene structure and parameters of a synthetic scene generator in order to produce good results. This process has many moving parts and is generally cumbersome to make work in a new application scenario. Doing so, such as learning the grammar itself, requires further investigation, and opens an exciting direction for future work.

\textbf{Acknowledgement:} We would like to acknowledge contributions through helpful discussions and technical support from Frank Cheng, Eric Cameracci, Marc Law, Aayush Prakash and Rev Lebaredian.
\section{Supplementary Material}

In the Supplementary Material, we present all experimental details in Sec.~\ref{sec:exp_deets} and 
show additional qualitative results in Sec.~\ref{sec:qual}.

\subsection{Experimental Details}
\label{sec:exp_deets}
\subsubsection{Loss gradient scale}
We estimate the log ratio $\ln \frac{q_f}{p_f}$ using kernel density estimators. The magnitude of this value can be large and lead to unstable training. We scale this value by $10^{-2}$. This was chosen empirically in order to match the magnitude of the MMD. Since we have $\lambda = 10^{-2}$ in our original loss and $r_\text{reject}(F)$ is independent of $F$ and dependent only on the structure generation model we rewrite the loss gradient as
\begin{align*}
\nabla_\theta \mathcal{L} \approx 10^{-2}\bigg( \frac{1}{M} \sum_{j = 1}^m (\ln \tilde{q}_f(\varphi(v_j')) -  \ln \tilde{p}_f(\varphi(v_j'))) \nabla_\theta \log q_I (v_j') + \mathbf{1}_{(1,\infty)}(r_\text{reject}) \bigg)
\end{align*} 
where $r_\text{reject}$ is the rejection rate (rejections per successful sample) of the model when sampling the $m$ images in the batch.
\subsubsection{Multi-MNIST}
\textbf{Feature Network Architecture:} We use a ResNet architecture consisting of 6 residual blocks and 3 fully connected layers. We use the standard residual block consisting of two 3x3 convolutions with batch normalization and leaky relu activations.  Average pooling is performed before passing the output to the fully connected layers. The final fully-connected layer outputs a vector of dimension 10 which corresponding to the logits for detecting whether a given digit class is in the scene. Since the Multi-MNIST images are grayscale, we duplicate them across channels to create a 3-channel image. The Multi-MNIST images are 256x256. The features used for training our structure generation model come from the average pool layer and the first two fully connected layers.

\textbf{Structure Generation Model Architecture:} The logits used for sampling are generated in an autoregressive fashion using an LSTM. The input of the LSTM is a one-hot encoding of the previous rule index (70 dimensional, in the Multi-MNIST case). We used a hidden dimension of 50 for the LSTM. The embedding of the one-hot encodings are computing using a fully connected layer. The output of the LSTM is produced by a fully connected layer which takes the hidden state of the LSTM as input and produces the logits used for rule sampling.

\textbf{Feature Network Training:} We first sample 5000 prior structures and their corresponding images, then train our feature network until we reach 75\% accuracy on digit detections (around 10 epochs). We use the sigmoid cross entropy loss, and use the ADAM optimizer with learning rate $\epsilon = 10^{-3}$, $\beta_1 = 0.9$ and $\beta_2=0.999$ with a batch size of 50.

\textbf{Structure Generation Model Pre-Training:} Using the 5000 prior structures generated for training the feature network, we train our structure generation model to minimize the negative log likelihood of the generated structures. This is done for 10 epochs using the ADAM optimizer with learning rate $\epsilon = 10^{-3}$, $\beta_1 = 0.9$ and $\beta_2=0.999$ with a batch size of 100.

\textbf{Structure Generation Model Training:} We train using 5000 target images. We use a batch size of $m = 500$ and $l = 500$ for both the generated and real images. We train for 20 epochs. We use the ADAM optimizer with a learning rate of $\epsilon = 10^{-4}$, $\beta_1 = 0.9$ and $\beta_2 = 0.999$. We also use a moving average baseline with $\alpha = 0.05$ in order to reduce the variance of the REINFORCE estimator.
\subsubsection{Aerial 2D}
\textbf{Feature Network Architecture:} We use the same feature network architecture as the Multi-MNIST with the final fully connected layer producing a vector of dimension 100. This vector is then split into 4 vectors of dimension 25 which correspond to the logits for one hot encodings. The one hot encodings represent (ranging from 0 to 24) represent the total number of objects of that class (cars, roads, trees and houses) in the scene. Aerial images are 256x256. The features used for training our structure generation model come from the average pool layer and the first two fully connected layers.

\textbf{Feature Network Training:} We first sample 5000 prior structures and their corresponding images, then train our feature network until we reach 75\% accuracy on car counting and road counting while ignoring the accuracy figures for trees and houses. Our loss function is the average of the cross entropy loss for each scene element. We use the ADAM optimizer with learning rate $\epsilon = 10^{-3}$, $\beta_1 = 0.9$ and $\beta_2=0.999$.

\textbf{Structure Generation Model Pre-Training:} Using the 5000 prior structures generated for training the feature network, we train our structure generation model to minimize the negative log likelihood of the generated structures. This is done for 10 epochs using the ADAM optimizer with learning rate $\epsilon = 10^{-3}$, $\beta_1 = 0.9$ and $\beta_2=0.999$ with a batch size of 100.

\textbf{Structure Generation Model Training:} We train using 5000 target images only (no structure ground truth). We use a batch size of $m = 500$ and $l = 500$ for both the generated and real images. We train for 20 epochs. We use the ADAM optimizer with a learning rate of $\epsilon = 10^{-4}$, $\beta_1 = 0.9$ and $\beta_2 = 0.999$. We also use a moving average baseline with $\alpha = 0.05$ in order to reduce the variance of the REINFORCE estimator.

\subsubsection{3D Driving Scenes}
\textbf{Structure Generation Model Architecture:} We use the same architecture as the previous experiments except we use 3 models in conjunction with three scenarios, as defined by ~\cite{sdr18}. Each scenario has a slightly different grammar, as described below. 

\textbf{Scenario Grammars:}
Different grammars are used for each scenario following SDR. The most general is the city scenario grammar
\begin{align*}
    \text{S}_\text{city} &\to \text{Street} \ \ \text{Outer}_L \ \ \text{Outer}_R\\
    \text{Street} &\to \text{Median} \ \text{Cars} \ | \ \text{Median} \  \text{Cars} \ \text{Cars} \ | \ \text{Cars} \ \text{Median} \ \text{Cars} \ |\\  
    &\ \text{Cars} \ \text{Median} \ \text{Cars} \ \text{Cars} \ |  \ \text{Cars} \ \text{Cars} \ \text{Median} \ \text{Cars} \ |\\ 
    &\ \text{Cars} \ \text{Cars} \ \text{Median} \ \text{Cars} \ \text{Cars}\\
    \text{Outer}_L &\to \text{Sidewalk} \ \text{Buildings} \ \text{Buildings} \ \text{Foliage}\\
    \text{Outer}_R &\to \text{Sidewalk} \ \text{Buildings} \ \text{Buildings} \ \text{Foliage}\\
    \text{Cars} &\to \text{car} \ | \  \epsilon \ | \  \text{Cars}\\
    \text{Foliage} &\to \text{tree} \ | \ \epsilon \ | \ \text{Foliage}\\
    \text{Buildings} &\to \text{building} \ | \ \epsilon \ | \ \text{Buildings}\\
    \text{Sidewalk} &\to \text{pole} \ | \ \text{sign} \ | \ \text{pedestrian} \ | \ \text{object} \ | \ \text{bike} \ | \ \epsilon \ | \ \text{Sidewalk}\\
    \text{Median} &\to \ \text{bush} \ | \ \text{object} \ | \ \text{tree} \ | \  \epsilon \ | \  M
\end{align*}
The suburban grammar is a restriction of the city grammar, specifically having one collection of buildings in the $\text{Outer}_L$ and $\text{Outer}_R$ non-terminals.
\begin{align*}
    \text{S}_\text{suburban} &\to \text{Street} \ \ \text{Outer}_L \ \ \text{Outer}_R\\
     \text{Street} &\to \text{Median} \ \text{Cars} \ | \ \text{Median} \  \text{Cars} \ \text{Cars} \ | \ \text{Cars} \ \text{Median} \ \text{Cars} \ |\\  
    &\ \text{Cars} \ \text{Median} \ \text{Cars} \ \text{Cars} \ |  \ \text{Cars} \ \text{Cars} \ \text{Median} \ \text{Cars} \ |\\ 
    &\ \text{Cars} \ \text{Cars} \ \text{Median} \ \text{Cars} \ \text{Cars}\\
    \text{Outer}_L &\to \text{Sidewalk} \ \text{Buildings} \ \text{Foliage}\\
    \text{Outer}_R &\to \text{Sidewalk} \ \text{Buildings} \ \text{Foliage}\\
    \text{Cars} &\to \text{car} \ | \  \epsilon \ | \  \text{Cars}\\
    \text{Foliage} &\to \text{tree} \ | \ \epsilon \ | \ \text{Foliage}\\
    \text{Buildings} &\to \text{building} \ | \ \epsilon \ | \ \text{Buildings}\\
    \text{Sidewalk} &\to \text{pole} \ | \ \text{sign} \ | \ \text{pedestrian} \ | \ \text{object} \ | \ \text{bike} \ | \ \epsilon \ | \ \text{Sidewalk}\\
    \text{Median} &\to \ \text{bush} \ | \ \text{object} \ | \ \text{tree} \ | \  \epsilon \ | \  M
\end{align*}
The rural grammar is a further restriction with the addition of a shoulder instead of a sidewalk
\begin{align*}
    \text{S}_\text{rural} &\to \text{Street} \ \ \text{Outer}_L \ \ \text{Outer}_R\\
    \text{Street} &\to  \text{Cars} \ \text{Median} \ \text{Cars} \ | \ \text{Cars} \ \text{Median} \ \text{Cars} \ \text{Cars} \ |\\
    &\ \text{Cars} \ \text{Cars} \ \text{Median} \ \text{Cars} \ | \ \text{Cars} \ \text{Cars} \ \text{Median} \ \text{Cars} \ \text{Cars}\\
    \text{Outer}_L &\to \text{Shoulder} \ \text{Foliage}\\
    \text{Outer}_R &\to \text{Shoulder} \ \text{Foliage}\\
    \text{Cars} &\to \text{car} \ | \  \epsilon \ | \  \text{Cars}\\
    \text{Foliage} &\to \text{tree} \ | \ \epsilon \ | \ \text{Foliage}\\
    \text{Median} &\to \epsilon
\end{align*}
The Median non-terminal represents a grass or stone median that divides the left and right portions of the street. Note that the order of non-terminals in a production rule does not always reflection position on the scene but rather the order in sampling. The position of a child object within its parent is decided by its parameters, that come from either a prior or are learnt, using the method of~\cite{kar2019metasim}.

\textbf{Feature Network Architecture:} We use the pool-3 layer of the inception-v3 network pre-trained on imagenet as our feature network.

\textbf{Structure Generation Model Pre-Training:} We sample 2000 prior structures for each scenario. We train the structure generation model of each scenario to minimize the negative log likelihood of the generated structures for the given scenario. This is done for 10 epochs using the default ADAM optimizer with a batch size of 100. We have two different priors for sampling structures. The prior from~\cite{kar2019metasim} and a simple prior. Both priors have the same sampling procedure but differ in terms of the distribution. In general sampling is done by sampling a random number of objects to be placed onto each container (the containers being Street, Outer, Shoulder, Sidewalk and Median).  The sampling is done using a uniform distribution bounded by $n_\text{min}$ and $n_\text{max}$ which determine the minimum and maximum number of objects in the container. In both priors the objects (i.e pole, sign, bike) are sampled with equal probability with replacement from the set of allowable objects in that specific container. An additional detail is determining the number of lanes on the road and whether or not the scene has a Median container and where should it be placed in the rural and urban scenarios. In both priors the number of lanes is sampled uniformly from 1 to 4 and the probability of a road having a median is 0.5. This can be interpreted as selecting a random production rule for the Street non terminal. Where the priors differ is in their choices of $(n_\text{min}, n_\text{max})$ for each container. In the prior from~\cite{kar2019metasim} the bounds are chosen to match the given scenario. For example in the rural setting there is a large bound on the number of foliage while in the urban setting it is smaller. Thus each container has a unique $(n_\text{min} n_\text{max})$ that is determined by the scenario. In the simple prior we restrict $n_\text{min} = 0$ and $n_\text{max} = 2$ for all containers in all scenarios.

\textbf{Structure Generation Model Training:}  We use a batch size of $l=300$ for the real images $V = \{v_1, \ldots, v_l\}$. The real images are taken from the train split of the KITTI dataset. To produce our generated batch we sample $m = 100$ images $B^s = \{v^s_1, \ldots, v^s_m\}$ for each scenario $s \in \{\text{city}, \text{suburban}, \text{rural}\}$. We then get KDE estimates
\begin{align*}
\tilde{p}_f(F) = \frac{1}{3 m} \sum_{s} \sum_{j = 1}^m K_{H}(F - \varphi(v_j^s)) && \tilde{q}_f(F) = \frac{1}{l} \sum_{j = 1}^l K_{H}(F - \varphi(v_j))
\end{align*}
The gradient of the loss for a single structure generation model is given by
\begin{align*}
\nabla_\theta \mathcal{L}^s \approx 10^{-2}\bigg( \frac{1}{M} \sum_{j = 1}^m (\ln \tilde{q}_f(\varphi(v_j^s)) -  \ln \tilde{p}_f(\varphi(v_j^s))) \nabla_\theta \log q_I (v_j^s) + \mathbf{1}_{(1,\infty)}(r_\text{reject}^s) \bigg)
\end{align*} 
so the densities are calculated using the whole generated set but only use images generated by that scenario for the REINFORCE sample. We use the ADAM optimizer with learning rate $\epsilon = 10^{-4}$, $\beta_1 = 0.9$ and $\beta_2 = 0.999$. We train for 5 epochs and use a moving average baseline with $\alpha = 0.05$ in order to reduce the variance of the REINFORCE estimator.

\textbf{Mask-RCNN Training:} We train a Mask-RCNN~\cite{he2017mask}\footnote{Using code from: https://github.com/facebookresearch/maskrcnn-benchmark} network on a generated training set and then test on the KITTI validation set~\cite{kitti}. In both the original and simple prior experiments we produce the generated image set by generating 10K samples for each experiment setting, with the scenario being chosen uniformly. For training, we render images with random saturation and contrast, which we observed to work better. Both the good prior and simple prior have 3 experiment settings which correspond to the rows of the Tab. 1 and 2. The training procedure for both experiments is the same.\\

First, we train on 10K samples from the probabilistic grammar. We use the ADAM optimizer with learning rate $\epsilon = 10^{-3}$, $\beta_1 = 0.9$ and $\beta_2 = 0.999$ with a batch size of 2. We finetune every row of Tab. 1 and Tab. 2 from the previous row. For the next two rounds (corresponding to the second and third rows of Tab. 1 and 2), we use learning rates of $5 * 10^{-4}$ and $10^{-4}$. 

\subsection{Additional Qualitative Results}
We show additonal generated images for the simple prior experiment in Fig.~\ref{fig:kitti_images}. Notice how the prior images (left) are mostly empty, while our learnt generated images have a structure that matches the real images much better. We notice that our method is restricted by the quality of the grammar, and therefore cannot generate structures such as parking side lanes (Row 5 KITTI Image) or intersubsections (Row 7.11 KITTI image) etc. With more work on writing a grammar, we hope to learn much better structures to cover more scenarios. Training for data generation is still very important after the effort on making the grammar, since for different downstream domains (such as driving in different countries), the target dataset can change drastically and appropriate synthetic data must be generated, where learning can help mitigate long cycles of careful tuning of scene parameters by humans. 

\label{sec:qual}
\begin{figure}[h!]
  {\hfill
  \includegraphics[width=0.32\textwidth]{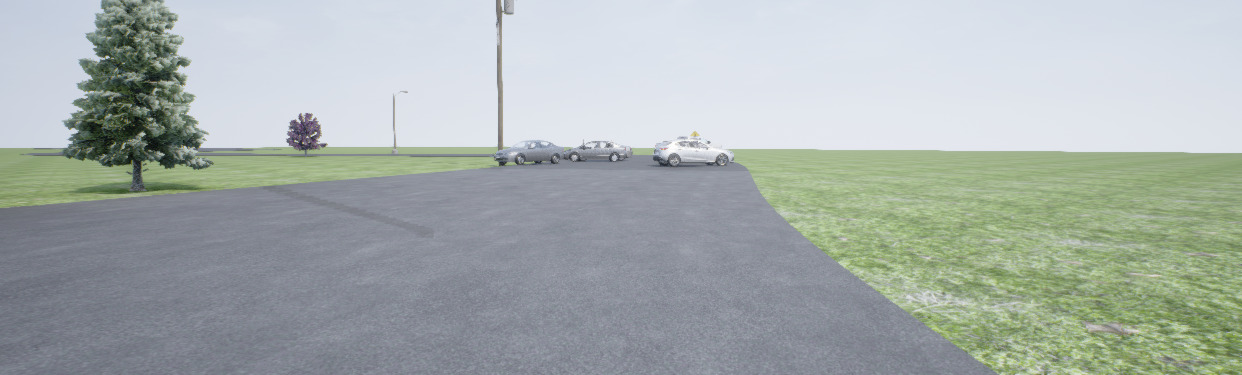}
  \hfill
  \includegraphics[width=0.32\textwidth]{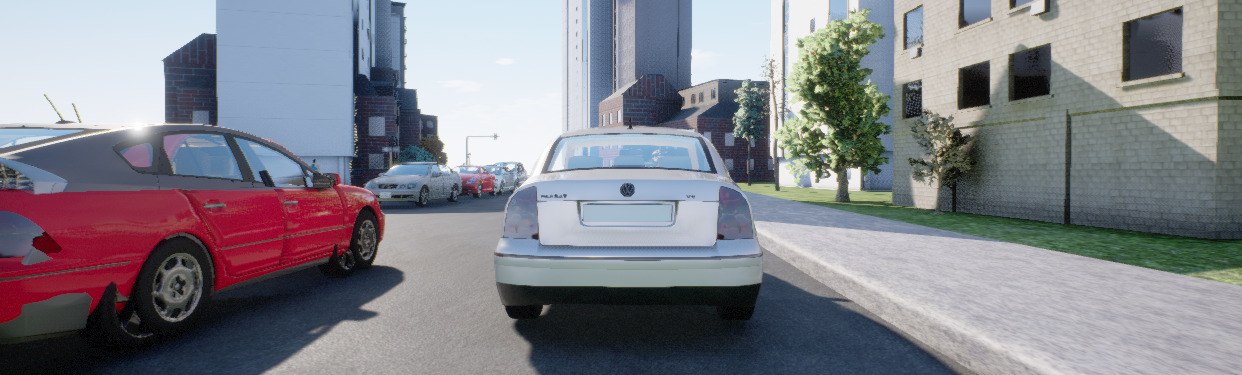}
  \hfill
  \includegraphics[width=0.32\textwidth]{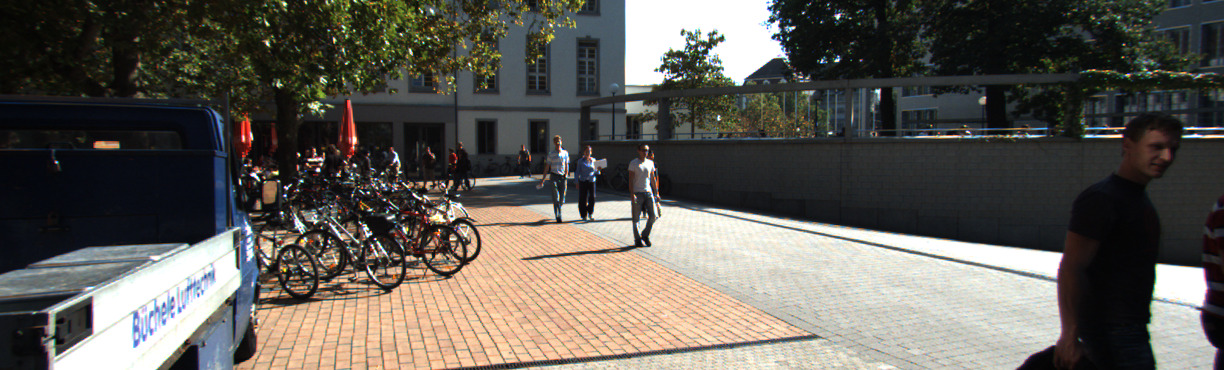}
  \hfill}
  
  {\hfill
  \includegraphics[width=0.32\textwidth]{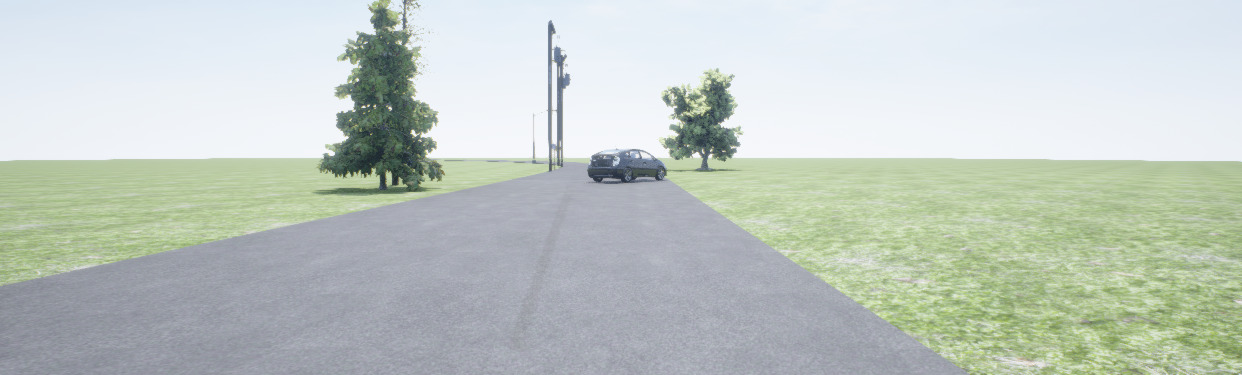}
  \hfill
  \includegraphics[width=0.32\textwidth]{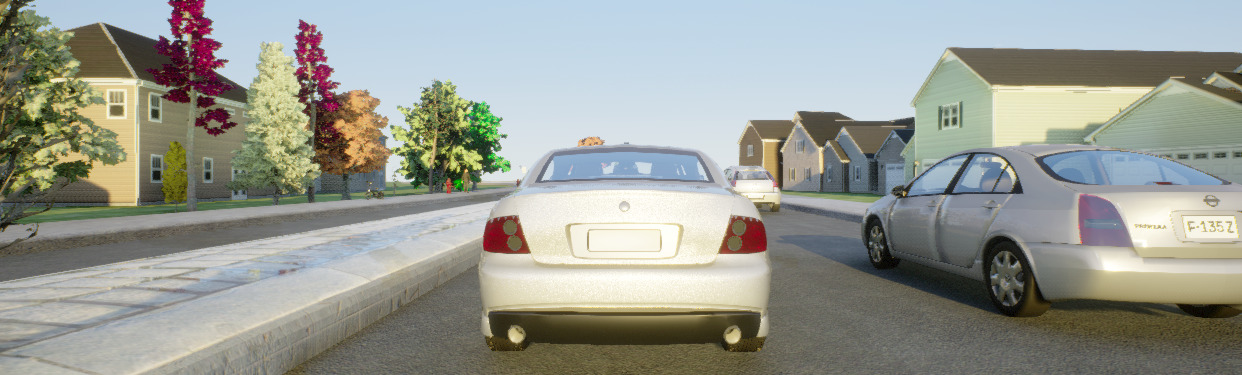}
  \hfill
  \includegraphics[width=0.32\textwidth]{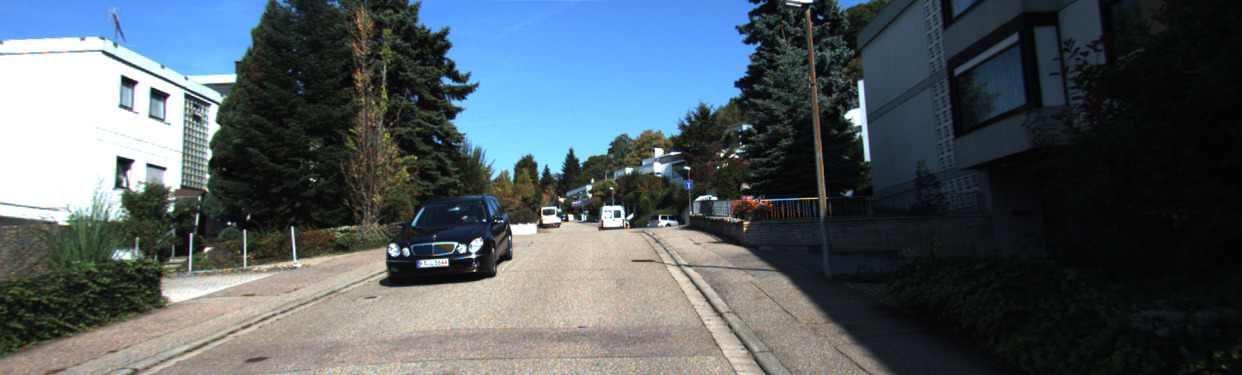}
  \hfill}

  {\hfill
  \includegraphics[width=0.32\textwidth]{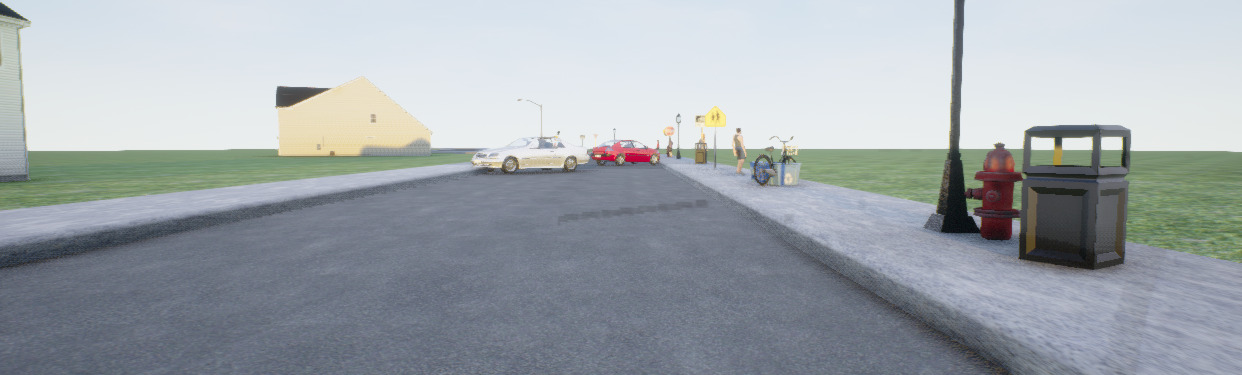}
  \hfill
  \includegraphics[width=0.32\textwidth]{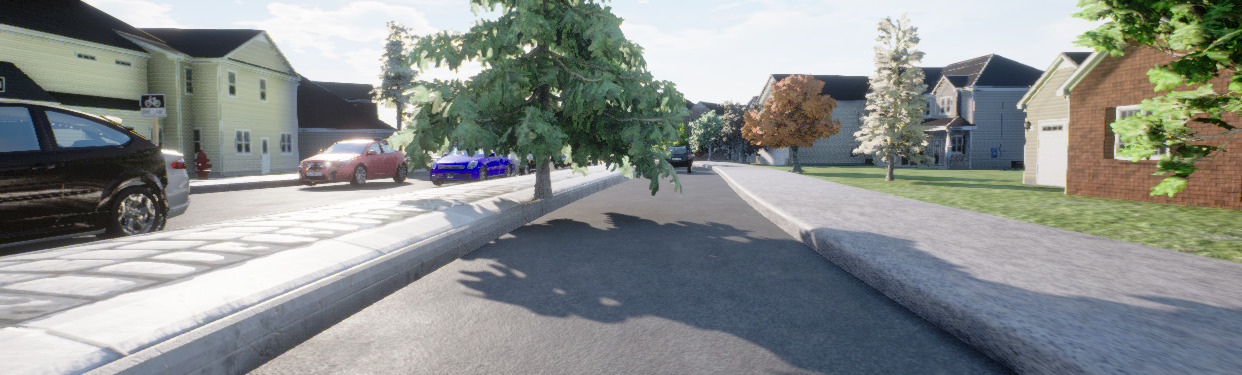}
  \hfill
  \includegraphics[width=0.32\textwidth]{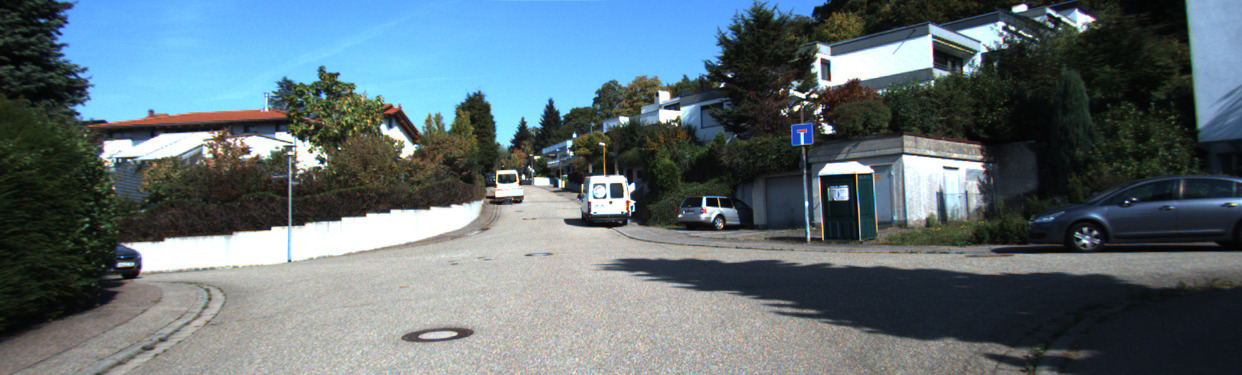}
  \hfill}

  {\hfill
  \includegraphics[width=0.32\textwidth]{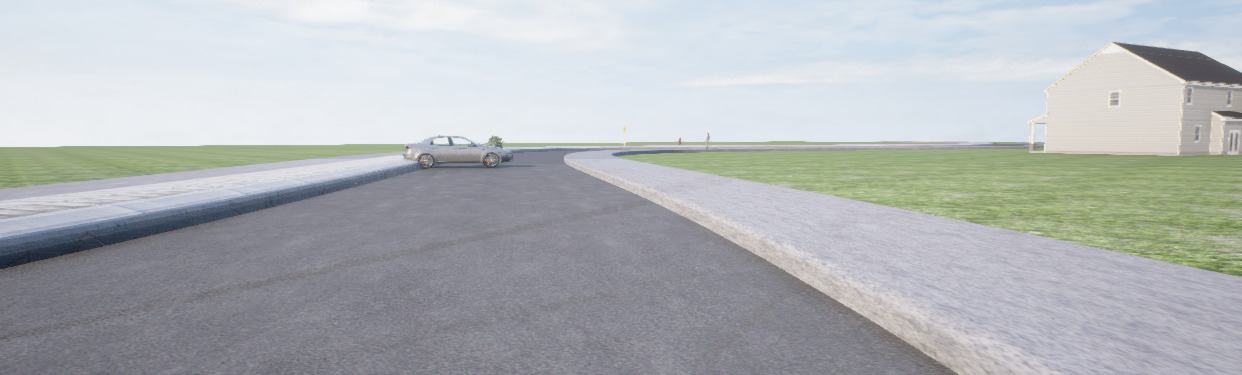}
  \hfill
  \includegraphics[width=0.32\textwidth]{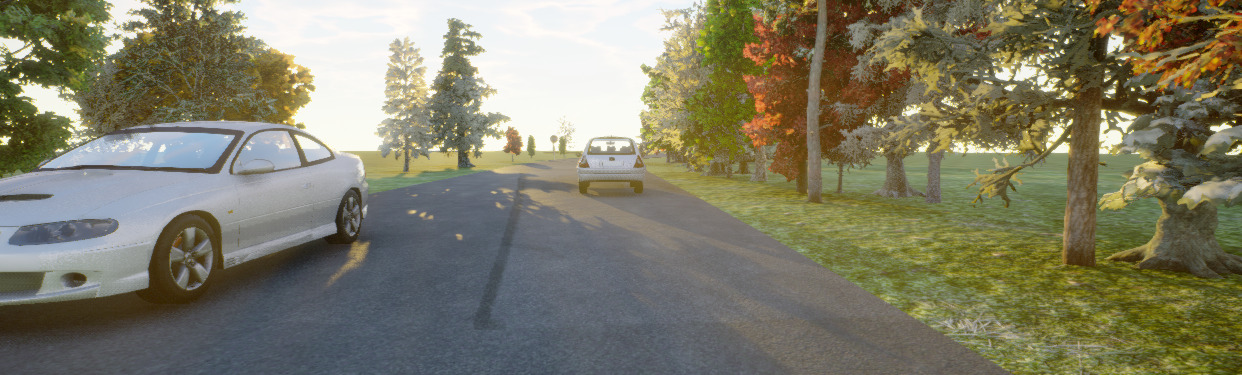}
  \hfill
  \includegraphics[width=0.32\textwidth]{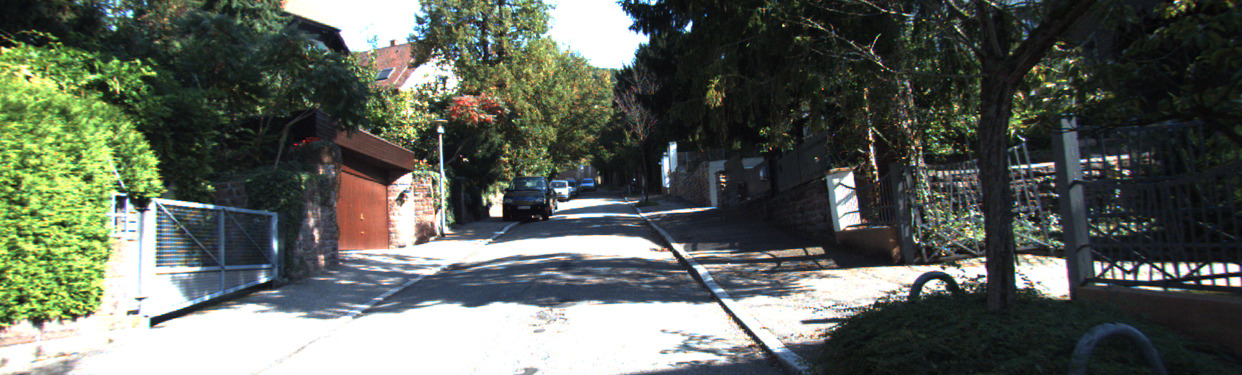}
  \hfill}

  {\hfill
  \includegraphics[width=0.32\textwidth]{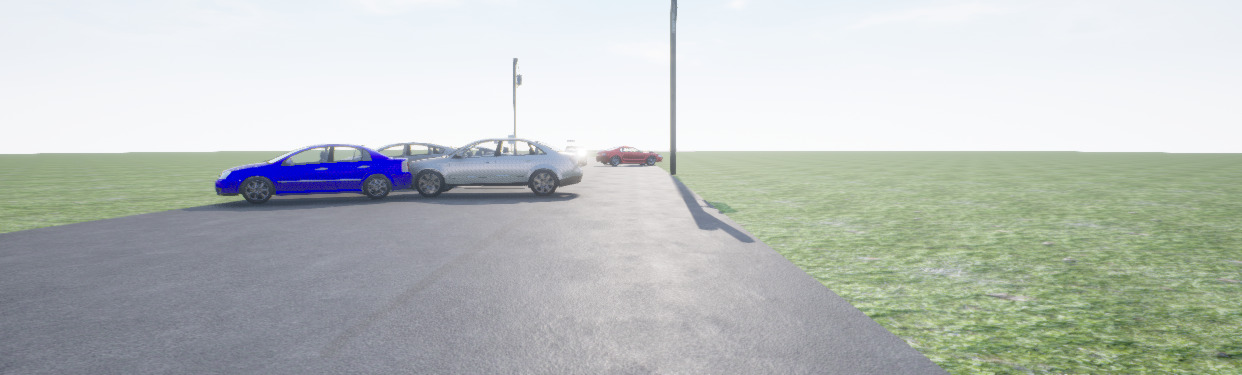}
  \hfill
  \includegraphics[width=0.32\textwidth]{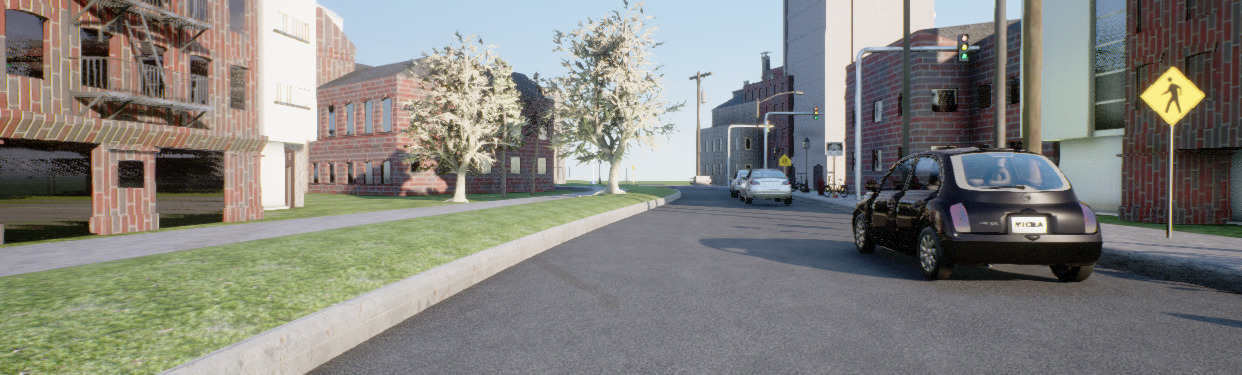}
  \hfill
  \includegraphics[width=0.32\textwidth]{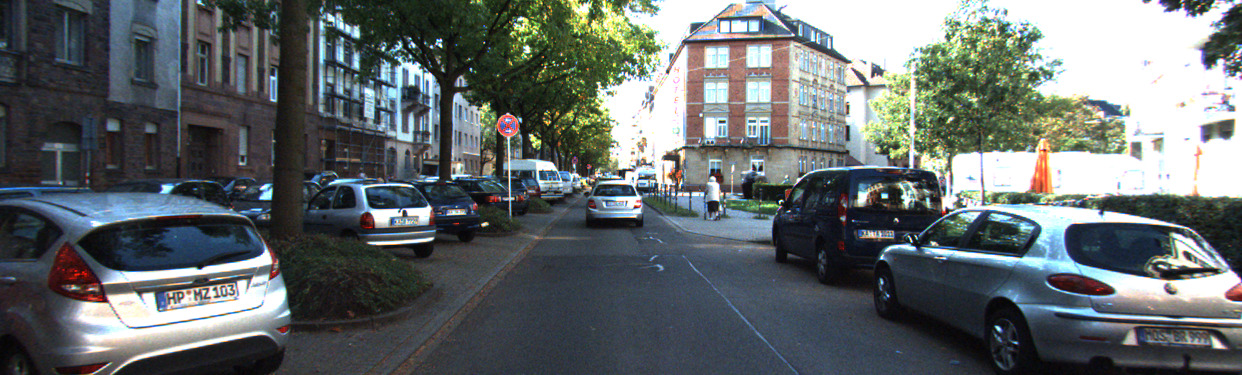}
  \hfill}

  {\hfill
  \includegraphics[width=0.32\textwidth]{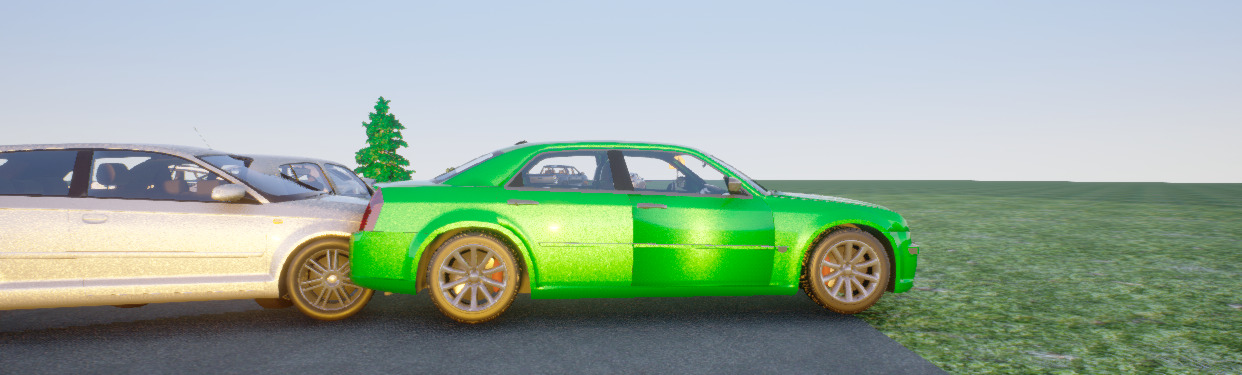}
  \hfill
  \includegraphics[width=0.32\textwidth]{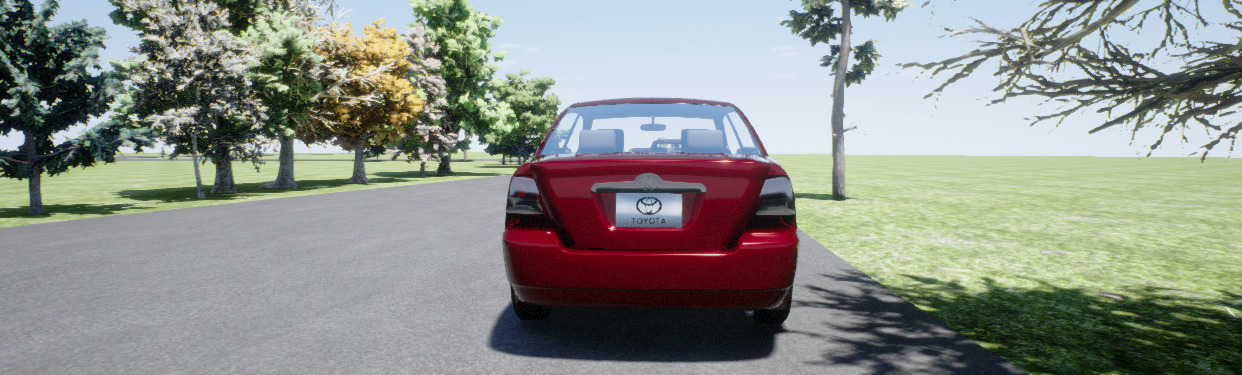}
  \hfill
  \includegraphics[width=0.32\textwidth]{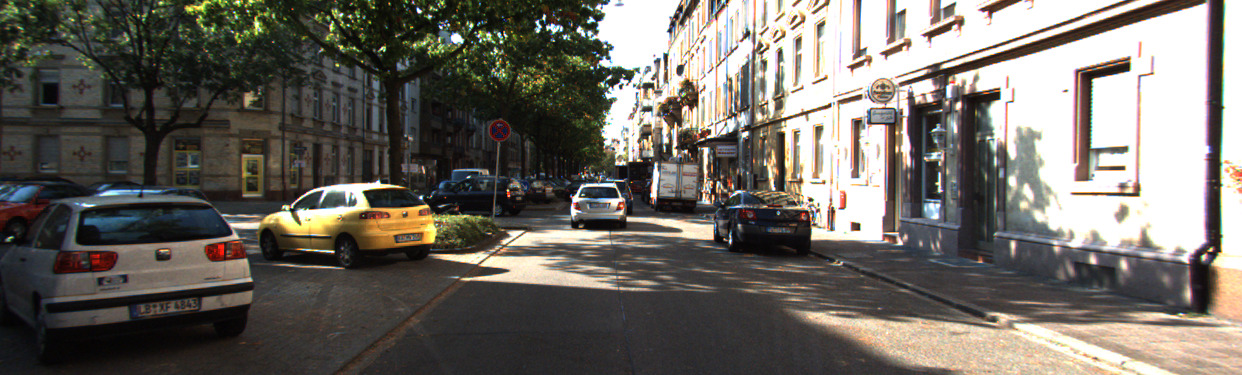}
  \hfill}

  {\hfill
  \includegraphics[width=0.32\textwidth]{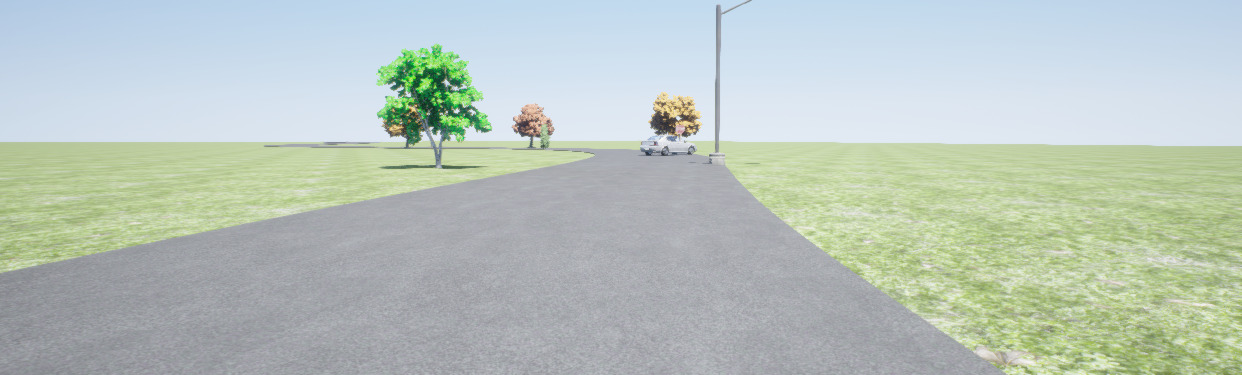}
  \hfill
  \includegraphics[width=0.32\textwidth]{SUPPL_TRAINED/000191.jpg}
  \hfill
  \includegraphics[width=0.32\textwidth]{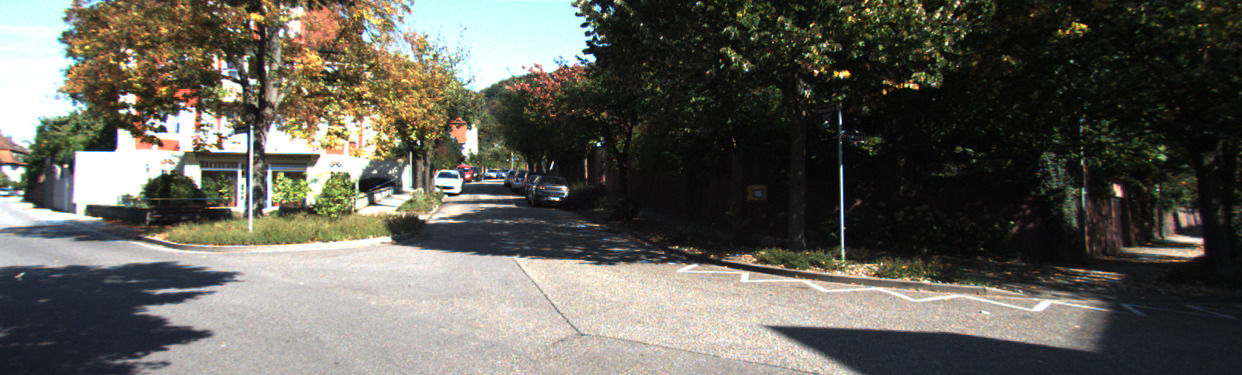}
  \hfill}

  {\hfill
  \includegraphics[width=0.32\textwidth]{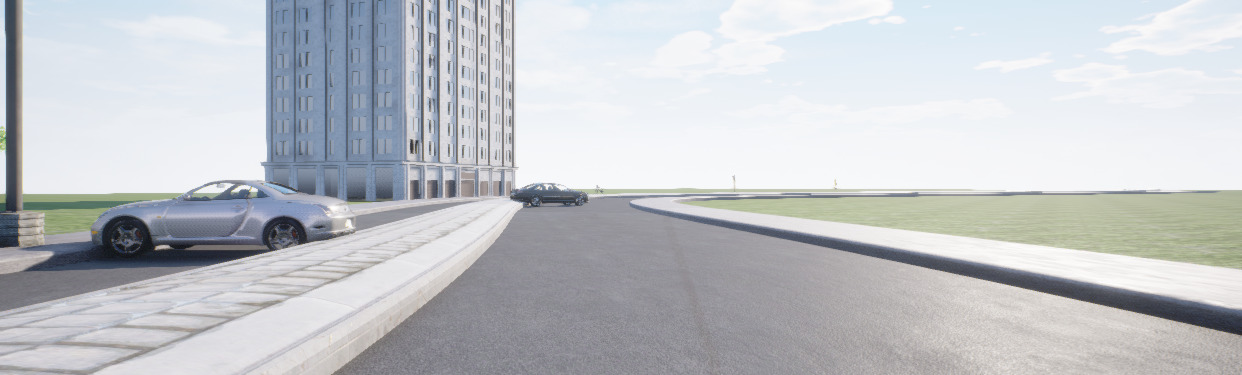}
  \hfill
  \includegraphics[width=0.32\textwidth]{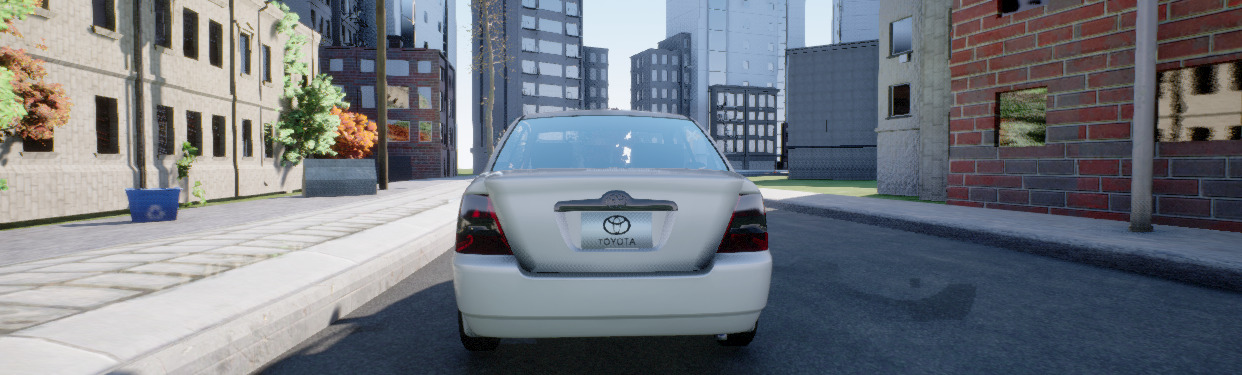}
  \hfill
  \includegraphics[width=0.32\textwidth]{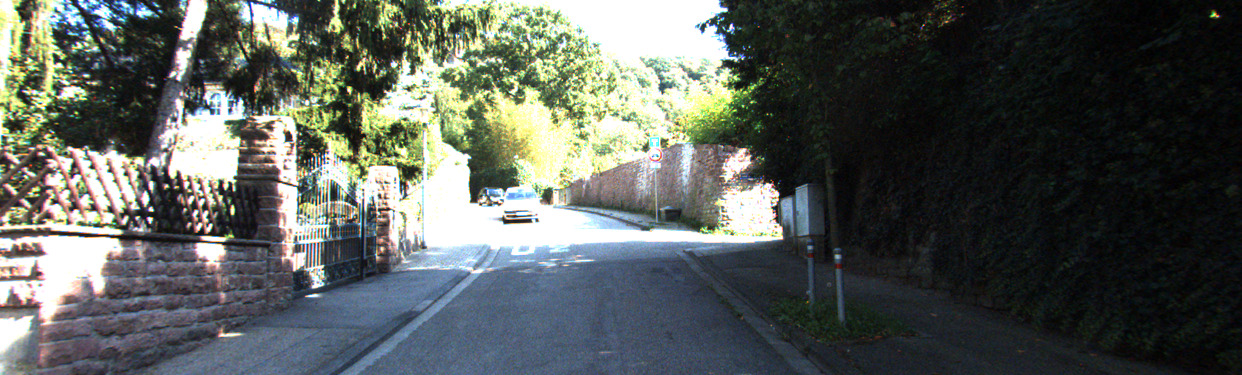}
  \hfill}

  {\hfill
  \includegraphics[width=0.32\textwidth]{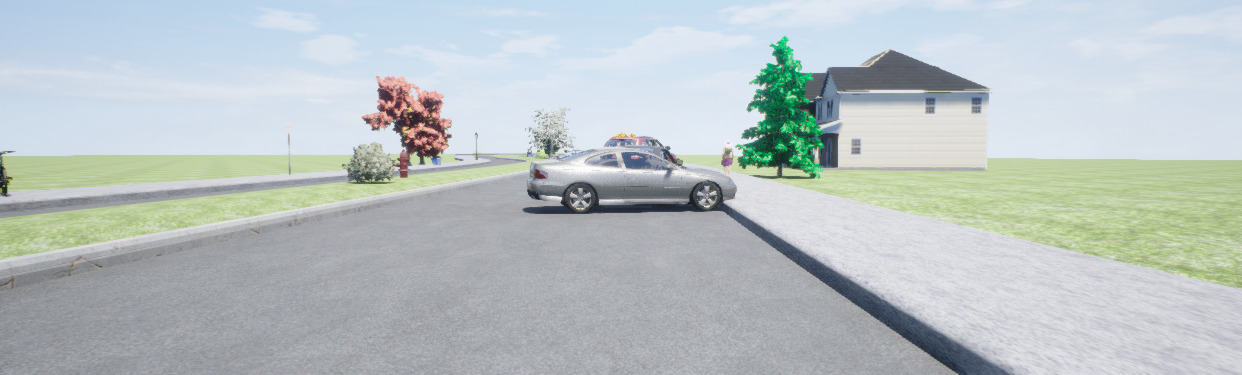}
  \hfill
  \includegraphics[width=0.32\textwidth]{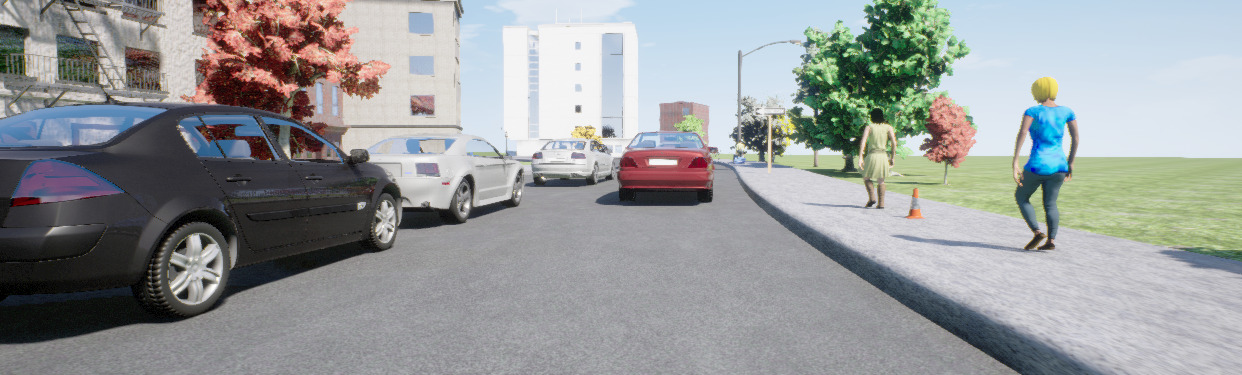}
  \hfill
  \includegraphics[width=0.32\textwidth]{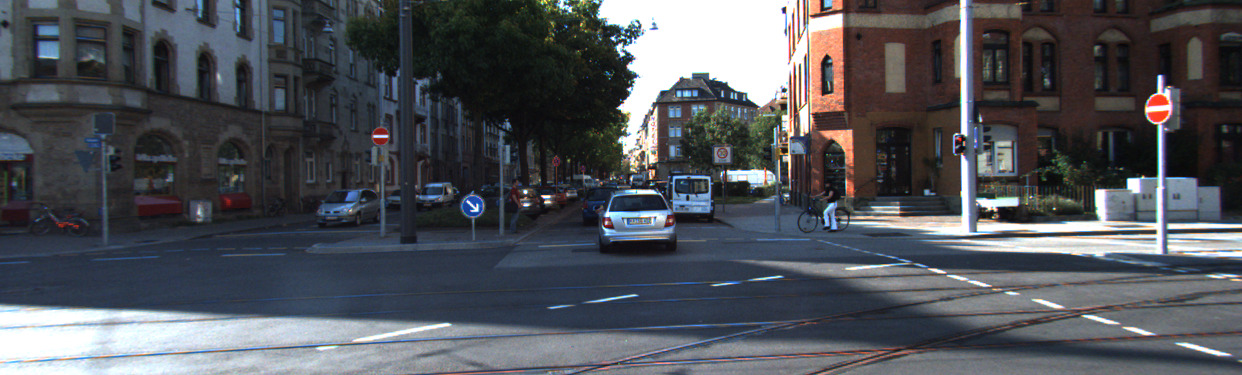}
  \hfill}

  {\hfill
  \includegraphics[width=0.32\textwidth]{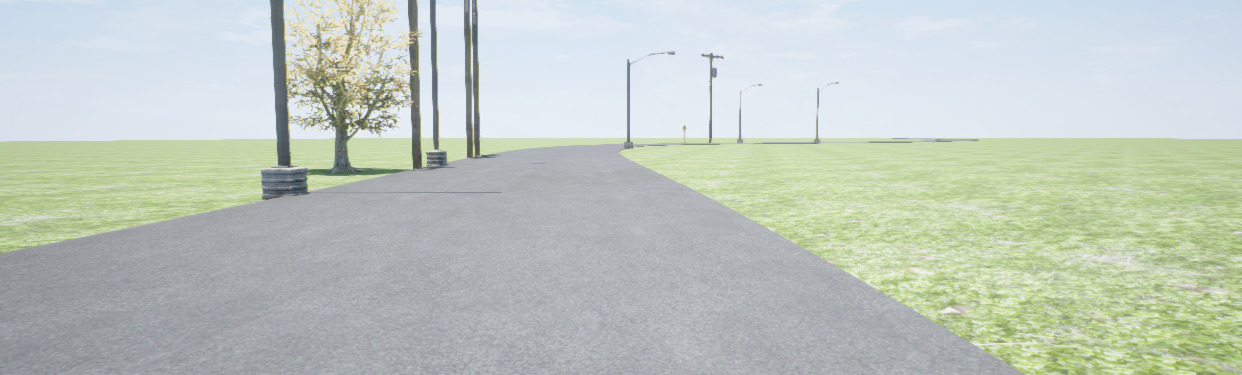}
  \hfill
  \includegraphics[width=0.32\textwidth]{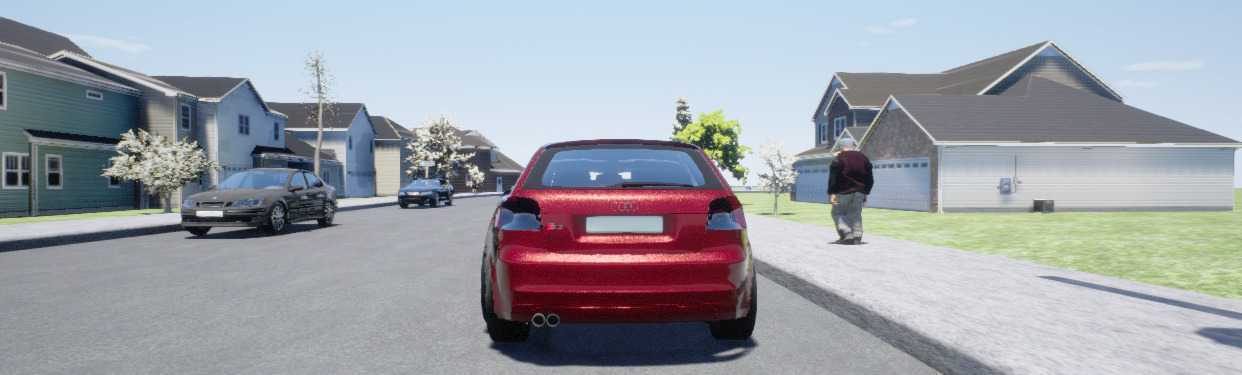}
  \hfill
  \includegraphics[width=0.32\textwidth]{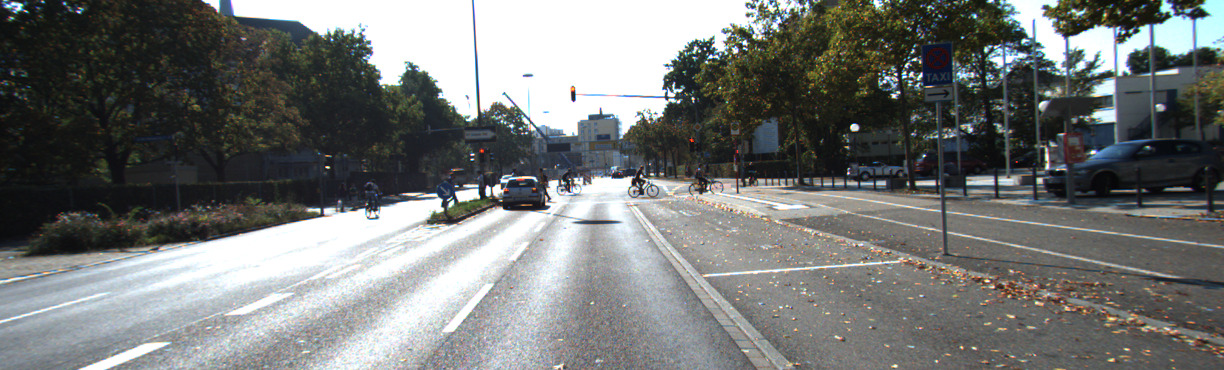}
  \hfill}

  {\hfill
  \includegraphics[width=0.32\textwidth]{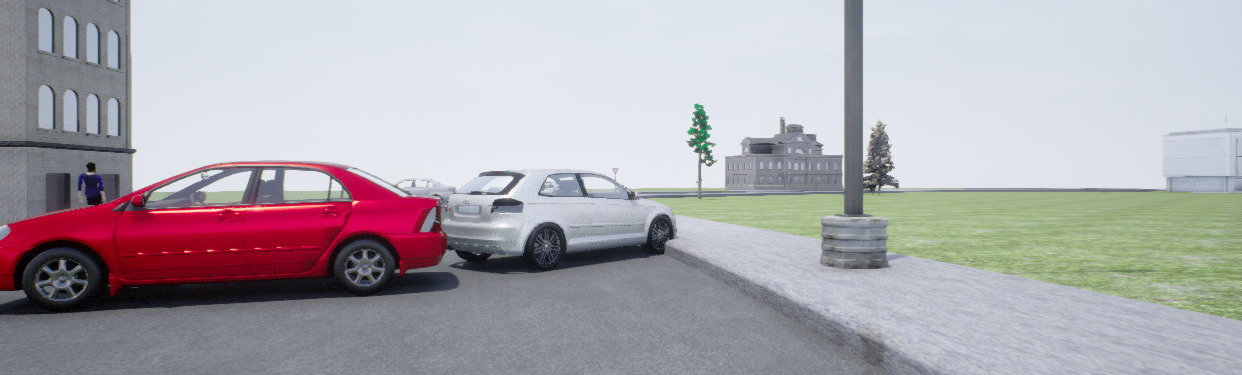}
  \hfill
  \includegraphics[width=0.32\textwidth]{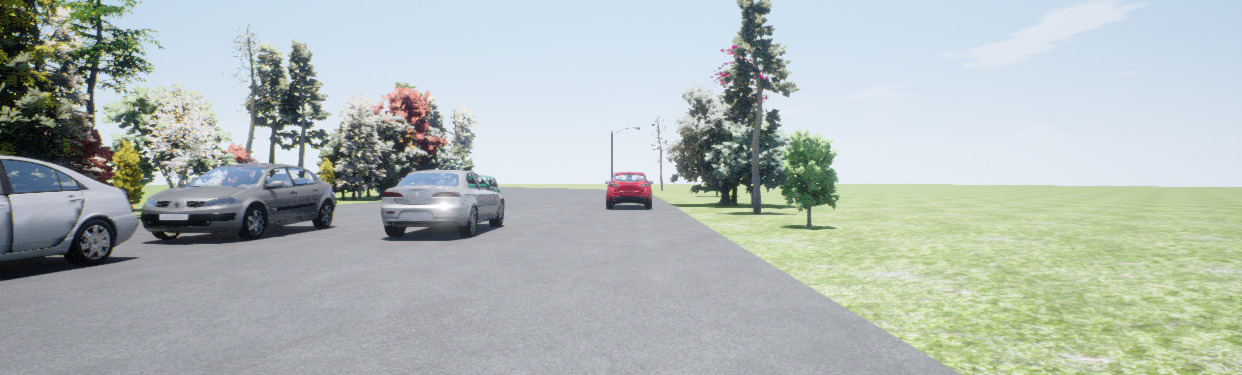}
  \hfill
  \includegraphics[width=0.32\textwidth]{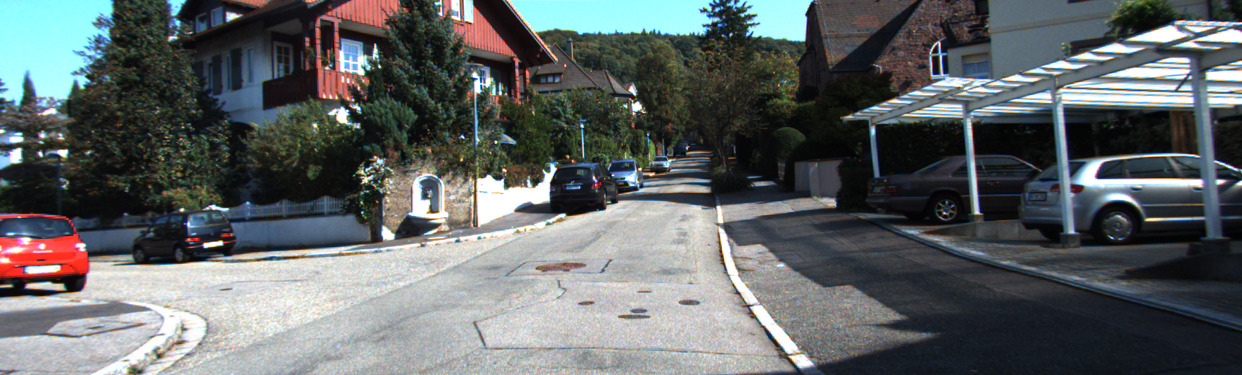}
  \hfill}

%   {\hfill
%   \includegraphics[width=0.32\textwidth]{SUPPL_BAD_PRIOR/000001.jpg}
%   \hfill
%   \includegraphics[width=0.32\textwidth]{SUPPL_TRAINED/000132.jpg}
%   \hfill}

  \caption{Random generated samples from the simple prior experiment. (Left) Using both the structure and parameter prior, (Middle) Using our learnt structure and parameters and (Right) random KITTI images Note: images in the same row are not correlated}
  \label{fig:kitti_images}
\end{figure}

\clearpage
\bibliographystyle{splncs04}
\bibliography{egbib}

%-------------------------------------------------------------------------

\bibliographystyle{splncs04}
\bibliography{egbib}

\end{document}